\documentclass[10pt,journal,compsoc]{IEEEtran}
\usepackage{cuted}
\usepackage{stackrel}
\usepackage{graphicx}
\usepackage{subcaption}
\usepackage{xfrac}
\usepackage{array}
\usepackage{stfloats}
\usepackage{multirow}
\usepackage{color}
\usepackage{tabulary}
\usepackage[usenames,dvipsnames]{xcolor}
\usepackage{colortbl}
\usepackage{pbox}
\usepackage{booktabs}
\usepackage{setspace}
\usepackage{bigints}
\usepackage{float}
\usepackage{url}
\usepackage[T1]{fontenc}
\usepackage{regexpatch}
\usepackage{amsmath,amssymb,amsthm,amsbsy,amsfonts}
\usepackage{cuted}
\usepackage{comment}
\usepackage{hyperref}
\usepackage{makecell}
\usepackage{algorithm}
\usepackage{algorithmic}

\ifCLASSOPTIONcompsoc
  \usepackage[nocompress]{cite}
\else
  \usepackage{cite}
\fi

\DeclareMathOperator{\sign}{sign}

\providecommand{\abs}[1]{\left\lvert#1\right\rvert}
\newcommand\norm[1]{\left\lVert#1\right\rVert}

\usepackage[normalem]{ulem} 

\providecommand{\abs}[1]{\lvert#1\rvert}
\providecommand{\norm}[1]{\lVert#1\rVert}

\newcommand{\subparagraph}{}

\usepackage{xcolor,cancel}
\usepackage{longtable}

\newcommand{\att}[1]{\textcolor{red}{#1}}

\begin{document}
\newcounter{MYtempeqncnt}
  \title{Multi-Task Model Personalization for Federated Supervised SVM in Heterogeneous Networks} 
 \author{Aleksei Ponomarenko-Timofeev, Olga Galinina, Ravikumar Balakrishnan, \\
 Nageen Himayat, Sergey Andreev, and Yevgeni Koucheryavy}

\IEEEtitleabstractindextext{%
\begin{abstract}
Federated systems enable collaborative training on highly heterogeneous data through model personalization, which can be facilitated by employing multi-task learning algorithms. However, significant variation in device computing capabilities may result in substantial degradation in the convergence rate of training. To accelerate the learning procedure for diverse participants in a multi-task federated setting, more efficient and robust methods need to be developed. In this paper, we design an efficient iterative distributed method based on the alternating direction method of multipliers (ADMM) for support vector machines (SVMs), which tackles federated classification and regression. The proposed method utilizes efficient computations and model exchange in a network of heterogeneous nodes and allows personalization of the learning model in the presence of non-i.i.d. data. To further enhance privacy, we introduce a random mask procedure that helps avoid data inversion. Finally, we analyze the impact of the proposed privacy mechanisms and participant hardware and data heterogeneity on the system performance.
\end{abstract}
}
\maketitle 

\IEEEdisplaynontitleabstractindextext

\IEEEraisesectionheading{\section{Introduction}\label{sec:introduction}}

\IEEEPARstart{T}{he} proliferation of connected smart devices has led to a surge in the volume of data available for machine learning (ML) in real-world applications. The cost of transmitting large volumes of data and privacy concerns~\cite{liu2021machine}  drive the development of federated learning (FL)~\cite{bonawitz2019towards,kairouz2021advances}, which offers a promising solution for distributed ML training across multiple devices without sharing their underlying data. 
FL opens numerous practical opportunities~\cite{ye2020edgefed}, such as learning models for scheduling tasks in a cellular network~\cite{ma2020scheduling,nishio2019client} or offloading computational tasks~\cite{yang2020computation}. 

In distributed scenarios, the devices may be located in different environments and collect statistically diverse data, which results in data heterogeneity. Conventional FL approaches may face challenges in producing a model, equally suitable for all participants. To enable personalization in scenarios with high data heterogeneity, the concept of multi-task learning (MTL)~\cite{caruana1997multitask} leverages the fact that models can be efficiently trained using a shared representation. 
The MTL paradigm has been widely applied, for example, to deep neural networks (DNN). Two methods for enabling MTL in the DNNs are soft and hard parameter sharing. Hard parameter sharing implies that certain task representation layers are synchronized while output layers are trained by each participant individually. This approach reduces the risk of overfitting, as shown in~\cite{baxter1997bayesian}. Contrary to hard parameter sharing, soft parameter sharing does not enforce the synchronization of layers but makes shared layers correlated. 

Although neural networks (NNs) provide powerful models, they are, in general, computationally expensive. One of their lightweight alternatives that can still achieve high accuracy in many problems while requiring fewer data~\cite{shao2012comparison}  is support vector machines (SVMs). In SVM, 
the supervised (i.e., classification or regression) problem is formulated as an optimization problem, which produces a hyperplane in a multi-dimensional space of features that characterize the object~\cite{kecman2005support}. The SVM-based methods essentially solve convex minimization problems, which are well-studied and may employ multiple numerical approaches. To efficiently utilize feature extraction, SVM-based solutions may rely on a set of tools available for data preprocessing ~\cite{jebara2004multi, christ2018time}.

Traditionally, distributed SVM problems~\cite{smith2017federated} are solved using conventional optimization techniques, such as the stochastic gradient descent (SGD) method, which might not be the most efficient choice in terms of performance in the case where multiple nodes iterate on disjoint sets of data. 
To accelerate the training procedure and further enhance the performance of distributed SVM-based solutions, we propose an MTL framework, which employs the alternating direction method of multipliers (ADMM) and allows for model personalization.
Particularly, we present a lightweight and robust multi-task algorithm for solving classification and regression problems in distributed systems  
of multiple nodes with personal datasets. We assess the impact of the heterogeneous computing capabilities of the participants.
Our contributions are presented as follows:

\begin{itemize}
    \item We present the mathematical formulation of the proposed MTL method for classification and regression problems. Based on this formulation, we propose an iterative algorithm that allows for model personalization in a system of multiple participants.
    \item We study the effects of the presence of stragglers, which are participants that are unable to complete a single epoch within a coordinator-set time limit due to poor computing capabilities. We assess the impact in the cases, where participants have equal computing capabilities as well as diverse capabilities.
    \item We design a masking mechanism, which provides privacy at the cost of distorting the content of the model update and explore its impact on the convergence of the algorithms.
    \item We compare the proposed algorithm with the state-of-the-art solution, which utilizes a similar approach but solves the primal problem, while our method solves the corresponding dual problem. We evaluate the performance in terms of conventional metrics and run-time of methods in a single-threaded mode.
\end{itemize}

The remainder of this paper is organized as follows. In Section~\ref{sec:rwork}, we overview state-of-the-art SVM-based solutions for distributed MTL. In Section~\ref{sec:pfor}, we present the mathematical formulation of the optimization problems. Sections~\ref{sec:mtc} and~\ref{sec:mtr} contain the derivation of iterative solutions for classification and regression problems, as well as the estimates of the run-time complexity. In Section~\ref{sec:sau}, we address data privacy and describe the proposed masking mechanism. Section~\ref{sec:numres} is dedicated to the simulated scenario description and discusses the observed behavior of the system. Finally, concluding remarks are provided in Section~\ref{sec:concl}.

\section{Related Work}
\label{sec:rwork}
To exploit the large volume of information available across multiple devices, numerous distributed learning solutions were developed~\cite{verbraeken2019survey}. If the models of different participants are correlated, the performance of the employed distributed method can be improved by applying MTL. 
For example, the authors of~\cite{al2020prediction} proposed an MTL framework for short-time traffic prediction using data collected from connected vehicles. Another MTL problem tackled by the authors of ~\cite{zhang2020app} is to predict the removal of applications from an online store based on the factors such as downloads and ratings. In this problem, MTL is required due to the fact that all of the applications are different (e.g., games and office suites), and thus gain and lose relevance according to dissimilar patterns. 

The field of distributed MTL has also received considerable attention, as highlighted in~\cite{zhang2021survey}. In particular, SVM, one of the conventional algorithms in the area of classification problems, was extended to accommodate distributed MTL formulations. An SVM-based framework for federated MTL in the presence of heterogeneous data was proposed in~\cite{smith2017federated}. Another approach based on SVM~\cite{10.1145/1014052.1014067} extended the single-task regularization-based method to MTL. For highly dynamic scenarios, where new features are introduced into the problem, the authors of~\cite{7899861} proposed an approach for transferring the knowledge from the model with an old set of features to a model with a new set of features. Further, the second-order cone programming (SOCP) to train multi-task learning support vector machines (MTL-SVMs) in terms of a primal problem was studied in~\cite{NIPS2007_67f7fb87}. 
A solution to decrease the number of solved tasks using SVMs while preserving the multi-task capability of the system was also explored by the authors of~\cite{multikernel}. Another approach to exploiting the interrelations between tasks was demonstrated by the authors of~\cite{marfoq2021federated}, where the distribution of data for each task is assumed to be a mixture of unknown distributions. 

Furthermore, with the growing concern for the privacy of user data in modern society~\cite{kairouz2019advances}, it becomes important to explore if the operational principle can provide privacy guarantees. While the existing works proposed SVM-based approaches to the problem of MTL, only a few focused on the asymptotic difficulty of the devised methods and data privacy analysis. Apart from privacy, the resilience of the framework towards adversaries, such as free-riders, who seek to obtain the model without participating in the training procedure, or curious servers, is a crucial aspect that should be addressed~\cite{lyu2020threats}. To tackle these threats, the authors of works~\cite{lv2021data} proposed strategies for estimating task relevance by calculating the correlation of the model update shared by a particular participant with the current global model. Authors of the study~\cite{huang2020exploratory} proposed a mechanism for scoring participants based on the produced shared model updates and analyzed various attack models. To discourage the devices from acting as free-riders, incentivization mechanisms can be applied~\cite{khan2020federated}. 
\par
In the case that no countermeasures are implemented in the system, a curious server may launch a membership inference attack~\cite{shokri2017membership}. Authors of work~\cite{ma2021privacy} devised a countermeasure for attacks such as a membership inference attack~\cite{liu2021machine}. To address the privacy problem, the authors of~\cite{shu2022clustered} proposed a dual-server architecture for the MTL framework. The framework also utilizes secure multi-party computation to enable secure mathematical operations and enforce the security of the data with CrypTen~\cite{knott2021crypten}. Other works explored the application of cryptography mechanisms such as multi-key fully homomorphic encryption (MK-FHE) to the MTL frameworks~\cite{stripelis2021secure}. While encryption allows participants to securely exchange even encrypted raw data, utilization of MK-FHE might be overly complex for devices with poor hardware capabilities. In such cases, differential privacy (DP) mechanisms can be applied~\cite{holohan2018bounded}, not only to SVM but also to DNN-based solutions~\cite{abadi2016deep}. 
%

\begin{figure}[ht!]
	\centering
	\includegraphics[width=0.85\linewidth]{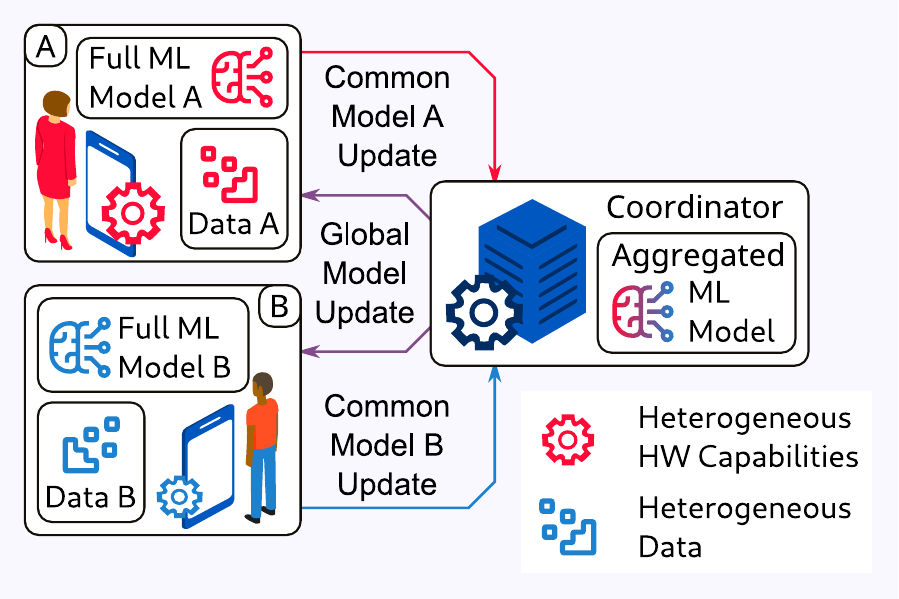}
	\caption{Envisioned distributed multi-task learning system.}
	\label{fig:physense}
\end{figure}
Prior works focused on distributed MTL, data and hardware heterogeneity, and privacy separately. Formulation of an efficient ML method and assessment of its performance under the heterogeneity conditions is a gap that we aim to bridge in this paper.

\section{Problem Formulation}
\label{sec:pfor}
In this section, we address the envisioned distributed multi-task learning scenario and define the corresponding soft-margin classification and regression problems. An example of the considered system is illustrated in Fig.~\ref{fig:physense}. Here, participants $A$ and $B$ with non-i.i.d. data iteratively train a common ML model. The third node, the coordinator, facilitates the training procedure by aggregating the model updates shared by $A$ and $B$ and responding with the result. The nodes may have different hardware, which leads to their diverse computing capabilities. 

Throughout further calculations and analysis, we represent vectors in bold, while scalars are given in regular font.

\subsection{General Scenario}
\label{subsec:gsce}
Our system consists of $K$ participants and one central coordinator (see Fig.~\ref{fig:physense}). Each participant $k$, $k=1,...,K$, holds a set of $n_k$ data samples $\pmb{x}_{ik} \in \mathbb{R}^{d}$, where $d$ is the dimensionality of the feature vector, and corresponding observations $y_{ik}$, $i=1,...,n_k$. The goal of each participant is to learn a decision function that recovers the observation from the provided data sample. In conventional SVM classification, the learned model is a set of parameters that define a \textit{separating hyperplane} for these classes. The \textit{decision function} can be defined as $a\left(\pmb{x}\right) = \pmb{w}^\intercal \pmb{x}$, where $\pmb{x}$ is a data point and $\pmb{w}$ is the separating hyperplane. 

We assume that the data are heterogeneous, i.e., their distribution varies from one participant to another. We define a \textit{task} as a problem of finding a participant-specific hyperplane defined by the properties of the data distribution at this participant. To tailor the learning model, we introduce a task-specific model component $\pmb{v}_k$ so that the decision function transforms to $a_k\left(\pmb{x}\right) = \pmb{w}^\intercal \pmb{x} + \pmb{v}^\intercal_k \pmb{x}$. 

The training process is iterative. When the participant with index $k$ takes part in collaborative learning at epoch $j$, it computes \textit{local} $\pmb{v}_k^{(j)}$ and \textit{global} $\tilde{\pmb{w}}_k^{(j)}$, which is shared with the coordinator. Then, the difference $\Delta \pmb{w}_k^{(j)}$ between the previous model, ${\pmb{w}}_k^{(j-1)}$, and updated $\tilde{\pmb{w}}_k^{(j)}$ is sent to the coordinator, which accumulates the differences from all participants for a fixed period of time $T_{\text{wait}}$. After the waiting period expires, the coordinator calculates global $\pmb{w}^{(j)}$ and returns the update to the participants. The participants run the next epoch of the local learning procedure using updated global $\pmb{w}^{(j)}$ and their individual local $\pmb{v}_{k}^{(j)}$. This process continues until the coordinator reaches the stopping criteria.



\begin{figure*}[t!]
	\centering
	\includegraphics[width=\linewidth]{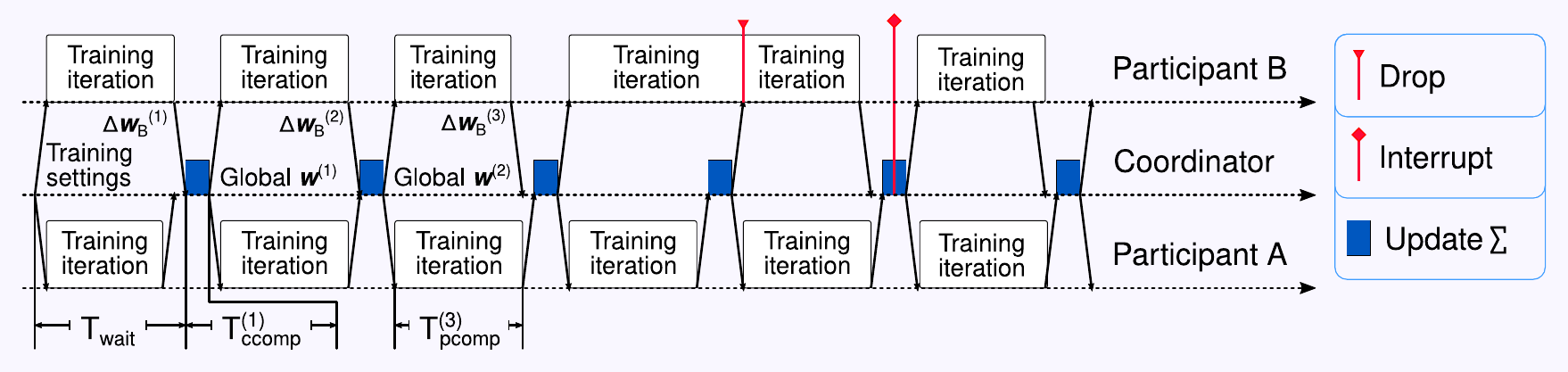}
	\caption{Message sequence chart of proposed training procedure.}
	\label{fig:MSC}
\end{figure*}

Furthermore, we introduce the random computing delay $T_{\text{pcomp},k}^{(j)}$ for participant $k$ at epoch $j$. Parameters of the delay distribution depend on both the number of data points, $n_k$, and the dimensionality, $d$, and their higher values result in a greater mean of the $T_{\text{pcomp},k}^{(j)}$. The coordinator is assumed to have a constant computing delay, $T_{\text{ccomp}}^{(j)}$, since it only computes the sum of the models sent by the participants.
If a participant fails to deliver its update to the coordinator within the waiting period, $T_\text{wait}^{(j)}$, it receives an update of the global hyperplane regardless. If the participant is in the process of calculating the local model while receiving an update from the coordinator, it then interrupts the ongoing calculations, applies the updated global model, and restarts the computation with current $\pmb{v}_k^{(j)}$. This enforces synchronization and prevents the accumulation of outdated updates if the participant struggles to complete the computations. 
We assume that the communication channel is sufficiently reliable and communications-related imperfections are out of the scope of this study. We also assume that communication delays are negligible compared to computation delays, thus, the updates are assumed to arrive instantly.
For an example of two participants, the message sequence chart is depicted in Fig.~\ref{fig:MSC}.
In the beginning, the coordinator sends the initial value of the global component, $\pmb{w}^{(0)}$, and other parameters. At epoch $i$, the participants compute their local and global model updates during a random time interval, $T_{comp}^{\left(i\right)}$.
Computing delays of different participants are not correlated since the hardware resources are not shared. After completing computations, the participants submit their global model updates $\Delta \pmb{w}^{(i)}_k$ to the coordinator for aggregating. The aggregation phase on the coordinator is more lightweight compared to the model calculation phase on participating participants. The corresponding summation time $T_{sum}$ is constant at each epoch. 
After the aggregation, the coordinator distributes the updated $\pmb{w}^{(0)}$ and continues accumulating updates during the time $T_{wait}$. If the participant receives an update from the coordinator, it interrupts the current calculations, applies the received global model, and restarts. If the coordinator receives an update from a participant during the summation operation, it continues the calculations and discards the belated update.

We further proceed to the problem formulation for MTL-SVM classification and regression with the notations provided in Table~\ref{tab:nottab}.

\begin{table}[h!]
	\centering
	\caption{Notation}
	\label{tab:nottab}
	\begin{tabular}{lr}
		\hline
		Notation & \makecell[c]{Description} \\
		\hline
		$\pmb{w}^{(j)}$ & Global model at epoch $j$\\
		$\Delta \pmb{w}^{(j)}$ & Global model update at epoch $j$\\
		$\Delta \pmb{w}^{(j)}_{k}$ & Participant $k$ shared model update at epoch $j$\\
		$\pmb{v}^{(j)}_{k}$ & Local model at participant $k$ at epoch $j$ \\
		$\pmb{x}_{ik}$ & $i$-th feature vector at participant $k$ \\
		$y_{ik}$ & $i$-th scalar label at participant $k$ \\
		$\xi_{ik}$ & $i$-th scalar slack variable at participant $k$ \\
		$\alpha_{ik}$ & $i$-th scalar Lagrange coefficient at participant $k$ \\
		$C_1$ & Slackness regularization constant \\
		$C_2$ & Task separation regularization constant \\
		$T_{wait}$ & Waiting period (set by coordinator) \\
		$T_{comp,k}^{(j)}$ & Computing delay at participant $k$ at epoch $j$\\
		\hline
	\end{tabular}
\end{table}

\subsection{Problem Formulation: Classification}
\label{subsec:pfc}
\par \textbf{Classification problem:} Given feature vectors $\pmb{x}_{ik} \in \mathbb{R}^{d}$ and labels $y_{ik} \in \{-1,1\}$ for $k = 1,...,T$, $i = 1,...,n_k$, find separating hyperplane components $\pmb{w}$ and $\pmb{v}_k$ for the participant-specific decision rule $a_k(\pmb x_{}) = \sign(\pmb{w}^\intercal \pmb{x}_{} + \pmb{v}_k^\intercal \pmb{x}_{})$. The optimal hyperplane can be found by solving the following \textit{primal} optimization problem:


\begin{equation}
	\begin{aligned}
		\min_{\pmb{w},\pmb{v}_k,} \quad & \frac{1}{2}\norm{\pmb{w}}^2 +\frac{C_2}{2} \sum \limits^{K}_{k=1}\norm{\pmb{v}_k}^2 \\
		\textrm{s.t.} \quad & y_{ik} \cdot \left( \pmb{w}^\intercal \pmb{x}_{ik} + \pmb{v}_k^\intercal \pmb{x}_{ik} \right) \geq 1, \\ 
		& i = 1,...,n_k, \quad k = 1,...,T,
	\end{aligned}
	\label{eq:prstrict}
\end{equation}
where $C_2$ is a regularization constant that controls the degree of similarity between the tasks. If $C_2$ is set to an infinitely large value, the solution would result in $\pmb{v}_k=0$, which corresponds to the single-task mode with one separating hyperplane for all participants. 

The two classes may be linearly inseparable, making primal problem~\eqref{eq:prstrict} infeasible. To address that, we utilize a soft margin SVM with the slack variables $\xi_{ik}$ and reformulate the primal task as follows:
\begin{equation}
	\begin{aligned}
		\min_{\pmb{w},\pmb{v}_k,\xi_{ik}} \quad & \frac{1}{2}\norm{\pmb{w}}^2 +\frac{C_2}{2} \sum \limits^{K}_{k=1}\norm{\pmb{v}_k}^2 + C_1  \sum \limits^{K}_{k=1} \sum \limits_{i=1}^{n_k} \xi_{ik}\\
		\textrm{s.t.} \quad & y_{ik} \cdot \left( \pmb{w}^\intercal \pmb{x}_{ik} + \pmb{v}_k^\intercal \pmb{x}_{ik} \right) \geq 1 - \xi_{ik},\\
		 & \xi_{ik} \geq  0, \quad i = 1,...,n_k, \quad k = 1,...,T,
	\end{aligned}
	\label{eq:prmsvm}
\end{equation}
where regularization constant $C_1$ controls the tolerance of the system towards outliers.

\subsection{Problem Formulation: Regression}
\label{subsec:pfr}
\par \textbf{Regression problem:} Given data $\pmb{x}_{ik} \in \mathbb{R}^{d}$ and labels $y_{ik} \in \mathbb{R}$ for $k = 1,...,T$, $i = 1,...,n_k$,
recover the (linear) relation between the dependent variable $y_{ik}$ and feature vector $\pmb{x}_{ik}$. For the participant $k$, the relation is defined by hyperplanes $\pmb{w}$ and $\pmb{v}_k$, which leads to \textit{decision function} $a_k(\pmb{x}_{}) = \pmb{w}^\intercal \pmb{x}_{} + \pmb{v}_k^\intercal \pmb{x}_{}$.

\begin{figure}[ht!]
	\centering
	\includegraphics[width=\linewidth]{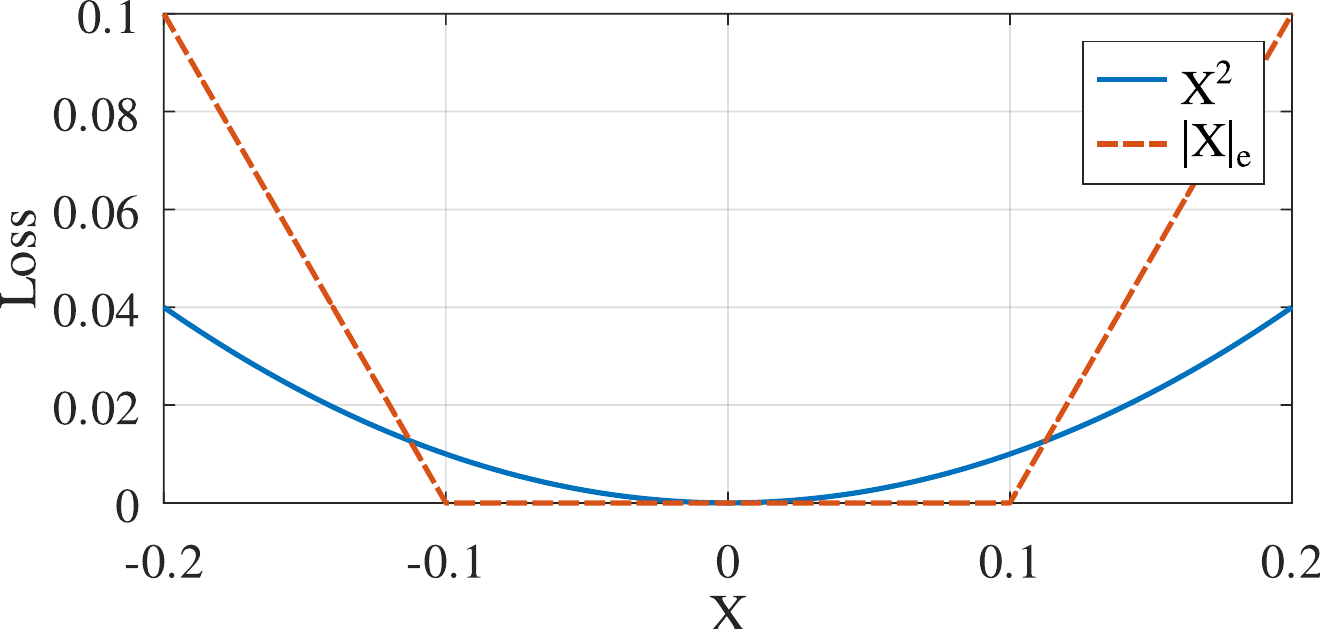}
	\caption{Modified loss function $|x|_{\epsilon}$ with $\epsilon=0.1$ tolerance compared to quadratic loss function.}
	\label{fig:lfunc}
\end{figure}

For the regression problem, we also employ the soft-margin formulation. To maintain the tractability of our method, we define the loss function as $Q(x) = \sum\limits_{k=1}^{K} \sum\limits_{i=1}^{n_k}  \abs{ \left(\pmb{w} + \pmb{v}_k\right)^\intercal \pmb{x}_{ik} - y_{ik} }_{\epsilon}$, where $\abs{z}_{\epsilon} = \max\left\{ 0, \abs{z}-\epsilon \right\}$. The selected loss function is illustrated in Fig.~\ref{fig:lfunc}. In this case, the slack variables correspond to the following two cases where the error caused by the noise is either above or below the hyperplane i.e.,
\begin{equation}
	\centering
	\begin{array}{cc}
		\xi_{ik}^+ = \left(a(\pmb{x}_{ik}) - y_{ik} - \epsilon\right), & \xi_{ik}^- = \left(-a(\pmb{x}_{ik}) + y_{ik} - \epsilon\right).
	\end{array}
	\label{eq:hilobound}
\end{equation}
We formulate the primal problem with the slack variables as
\begin{equation}
	\begin{aligned}
		\min_{\pmb{w},\pmb{v}_k ,\xi_{ik}^{+},\xi_{ik}^{-}} \quad & \frac{1}{2} \norm{\pmb{w}}^2 +  \frac{C_2}{2}  \sum \limits_{k=1}^{K}  \norm{\pmb{v}_k}^2 + C_1  \sum \limits_{k=1}^{K} \sum \limits_{i=1}^{n_k} \left(\xi_{ik}^{+}  + \xi_{ik}^{-}  \right)\\
	    \textrm{s.t.} \quad & \!\!\! \pmb{w}^\intercal \pmb{x}_{ik}  \geq  y_{ik}- \epsilon - \xi_{ik}^{-} , i = 1,...,n_k, t = 1, ..., T,\\
		& \!\!\! \pmb{w}^\intercal \pmb{x}_{ik}  \leq  y_{ik} + \epsilon + \xi_{ik}^{+}, i = 1,...,n_k, t = 1, ..., T,\\
		& \!\!\! \xi_{ik}^{-} \geq 0, \xi_{ik}^{+} \geq 0, i = 1,...,n_k, t = 1, ..., T.
	\end{aligned}
	\label{eq:regpr}
\end{equation}
In the following sections, we translate the primal problems for classification and regression into the corresponding dual formulations and solve them.


\section{Multi-task Classification}
\label{sec:mtc}
In this section, we formulate the dual problem for~\eqref{eq:prmsvm} and derive an iterative solution for participants in the multi-task setting.
\subsection{Dual Problem for Classification}
\label{subsec:dpc}
\par Let us convert the primal problem~\eqref{eq:prmsvm} into the dual problem using the Lagrangian $L(\pmb{w}, \pmb{v}_k, \pmb \xi)$ function
\begin{equation}
	\begin{array}{l}
		L(\pmb{w}, \pmb{v}_k, \pmb \xi) = \\ \frac{1}{2}\norm{\pmb{w}}^2
		+  \frac{C_2}{2} \sum \limits^{K}_{k=1}\norm{\pmb{v}_k}^2 + C_1  \sum \limits^{K}_{k=1} \sum \limits_{i=1}^{n_k} \xi_{ik}
		\!-\! \!\sum \limits_{k=1}^{K} \! \sum \limits_{i=1}^{n_k}  \eta_{ik} \xi_{ik} \\
		\!-\! \sum \limits_{k=1}^{K} \! \sum \limits_{i=1}^{n_k}   \alpha_{ik} \! \left[y_{ik}  \!\left( \pmb{w}^\intercal \pmb{x}_{ik} \!+\! \pmb{v}_k^\intercal \pmb{x}_{ik} \right) \!-\!1\! + \!\xi_{ik} \right]\!.
	\end{array}
	\label{eq:lagrmsvm}
\end{equation}
Here, $\alpha_{ik}$ and $\eta_{ik}$ are Lagrangian coefficients, 
which correspond to constraints $y_{ik} \cdot \left(\pmb{w}^\intercal \pmb{x}_{ik} + \pmb{v}_k^\intercal \pmb{x}_{ik}\right) \geq 1- \xi_{ik}$ and $\xi_{ik} \geq 0$, accordingly.
We reduce the number of variables in the Lagrangian by using the requirements for a saddle point existence. For a saddle point to exist, the following should hold: $\frac{\partial L}{\partial \pmb{w}} = 0$, $\frac{\partial L}{\partial \pmb{v}_k}  = 0$, and $\frac{\partial L}{\partial \xi_{ik}}  = 0$. Using these conditions, we may then express $\pmb{w}$, $\pmb{v}_k$, and $C_1$ as follows:
\begin{equation}
	\left\{ \!\!
	\begin{array}{l}
		 \frac{\partial L}{\partial \pmb{w}} = \pmb{w}  -\sum \limits_{k=1}^{K} \sum \limits_{i=1}^{n_k}  \alpha_{ik} y_{ik} \pmb{x}_{ik}, \\
		 \frac{\partial L}{\partial \pmb{v}_k} = C_2 \pmb{v}_k
		\!-\! \sum \limits_{i=1}^{n_k} \alpha_{ik} y_{ik}  \pmb x_{ik}, \\
		 \frac{\partial L}{\partial \xi_{ik}} = C_1  -  \eta_{ik} -\alpha_{ik},
	\end{array}
	\right. \!\!
	\Rightarrow
	\left\{  \!\!
	\begin{array}{l}
		\pmb{w} = \sum \limits_{k=1}^{K}\sum \limits_{i=1}^{n_k}  \alpha_{ik} y_{ik} \pmb{x}_{ik}, \\
		\pmb{v}_k=\frac{1}{C_2}\sum \limits_{i=1}^{n_k} \alpha_{ik} y_{ik}  \pmb{x}_{ik},  \\
		C_1  =  \eta_{ik} +\alpha_{ik}.
	\end{array}
	\right.
	\label{eq:derres}
\end{equation}

We substitute variables~\eqref{eq:derres} into the Lagrangian~\eqref{eq:lagrmsvm}. The arguments of $L(\pmb{w}, \pmb{v}_k, \xi)$ are reduced to variables $\alpha_{ik}$; $\pmb{x}_{ik}$ and $y_{ik}$ are fixed terms. For convenience, we solve the minimization problem $\min_{\alpha_{ik}}{\left(-L\left(\pmb\alpha\right)\right)}$, where $-L(\pmb\alpha)$ is defined as
\begin{equation}
	\begin{array}{l}
    \!\!-L(\pmb \alpha)\! = \\ \!\frac{1}{2}\norm{\sum \limits_{k=1}^{K}\!\sum \limits_{i=1}^{n_k} \! \alpha_{ik} y_{ik} \pmb{x}_{ik}}^2 
	\!\!\!+\!  \frac{1}{2C_2}\! \sum \limits^{K}_{k=1} \! \norm{\sum\limits_{i=1}^{n_k} \!  \alpha_{ik} y_{ik} \pmb{x}_{ik}}^2 
	\!\!\!-\!\! \sum \limits_{k=1}^{K} \!\sum \limits_{i=1}^{n_k} \!  \alpha_{ik}.
%
	\end{array}
	\label{eq:dualproblag}
\end{equation}
Hence, we arrive at the following dual optimization problem:
\begin{equation}
	\begin{array}{rl}
    \underset{\alpha_{ik}}{\min} \quad & \!\frac{1}{2}\norm{\sum \limits_{k=1}^{K}\!\sum \limits_{i=1}^{n_k} \! \alpha_{ik} y_{ik} \pmb{x}_{ik}}^2 \\
    & + \frac{1}{2C_2}\! \sum \limits^{K}_{k=1} \! \norm{\sum\limits_{i=1}^{n_k} \!  \alpha_{ik} y_{ik} \pmb{x}_{ik}}^2 
	\!\!\!-\!\! \sum \limits_{k=1}^{K} \!\sum \limits_{i=1}^{n_k} \!  \alpha_{ik} \\
	  \textrm{s.t.} \quad & 0\leq  \alpha_{ik} \leq  C_1, i = 1,...,n	. 
%
	\end{array}
	\label{eq:dualprob}
\end{equation}

We aim to solve problem~\eqref{eq:dualprob} by using ADMM~\cite{boyd2011distributed} and minimize $-L(\pmb \alpha)$ over each $\alpha_{ik}$ individually. We define an impact $z_{ik}(\alpha_{ik})$ of each $\alpha_{ik}$ on the value of the objective function as given below:
\begin{equation}
	\begin{array}{l}
	   z_{ik}(\alpha_{ik}) = \\
    \alpha_{ik} y_{ik} \pmb{x}_{ik}^\intercal \left( \sum \limits_{k_2=1}^{K}\sum \limits_{i_2=1}^{n_k} \alpha_{i_2k_2} y_{i_2k_2} {\pmb{x}_{i_2k_2}} - \alpha_{ik} y_{ik} \pmb{x}_{ik} \right) \\
+ \frac{1}{2} \alpha_{ik}^2 y_{ik}^2  \norm{\pmb{x}_{ik}}^2 
+  \frac{1}{C_2}  \alpha_{ik} y_{ik} \pmb{x}_{ik}^\intercal \sum\limits_{i_2=1, i_2 \neq i}^{n_k} \alpha_{i_2k} y_{i_2k} {\pmb{x}_{i_2k}} \\
+ \frac{1}{2C_2} \alpha_{ik}^2 y_{ik}^2 \norm{\pmb{x}_{ik}}^2 -  \alpha_{ik} .
	\end{array}
	\label{eq:zin}
\end{equation}
Our goal is to minimize each $z_{ik}(\alpha_{ik})$ sequentially. At each participant $k$, the index $i$ is selected randomly. 
Importantly, $z_{ik}(\alpha_{ik})$ is a quadratic function, and, therefore, by using condition $\frac{\partial z_{ik}(\alpha_{ik})}{\partial \alpha_{ik}} = 0$, we can obtain $\hat{\alpha}_{ik}$, at which the impact is at maximum. The condition on the derivative leads to the equation
\begin{equation}
	\begin{array}{l}
-1 + y_{ik} \pmb{x}_{ik}^\intercal \left( \sum \limits_{k_2=1}^{K}\sum \limits_{i_2=1}^{n_k} \alpha_{i_2k_2} y_{i_2k_2} {\pmb{x}_{i_2k_2}} - \alpha_{ik} y_{ik} \pmb{x}_{ik} \right) \\
+ \alpha_{ik} y_{ik}^2  \norm{\pmb{x}_{ik}}^2
+  \frac{1}{C_2}   y_{ik} {\pmb{x}_{ik}^\intercal} \sum\limits_{i_2=1, i_2 \neq i}^{n_k} \alpha_{i_2k} y_{i_2k} {\pmb{x}_{i_2k}} \\
+ \frac{1}{C_2} \alpha_{ik} y_{ik}^2 \norm{\pmb{x}_{ik}}^2 = 0.
	\end{array}
	\label{eq:alpha_eq}
\end{equation}
To compute the new value of $\alpha_{ik}$, we apply expressions for $\pmb{w}$ and $\pmb{v}_{k}$ in~\eqref{eq:derres}. At the start of each epoch, participants utilize current values of $\pmb{w}$ and $\pmb{v}_{k}$, which contain previously computed $\alpha_{ik}$ denoted as $\alpha_{ik}^{\text{(prev)}}$. Thus, we may use $\pmb{w}$ and $\pmb{v}_{k}$ as a link between two consecutive iterations and express parts of~\eqref{eq:alpha_eq} as provided below:
\begin{equation}
	\!\left\{ \!\!\!
	\begin{array}{l}
		\sum \limits_{k_2=1}^{K}\sum \limits_{i_2=1}^{n_k} \alpha_{i_2k_2} y_{i_2k_2} {\pmb{x}_{i_2k_2}} - \alpha_{ik} y_{ik} \pmb{x}_{ik}  = \pmb w -  \alpha_{ik}^{(\text{prev})}  y_{ik}  \pmb{x}_{ik},\!\! \\
    \frac{1}{C_2}  \sum\limits_{\substack{i_2=1 \\ i_2 \neq i}}^{n_k} \alpha_{i_2k} y_{i_2k} {\pmb{x}_{i_2k}} = \pmb{v}_k -  \frac{1}{C_2}\alpha_{ik}^{(\text{prev})} y_{ik} {\pmb{x}_{ik}}.
	\end{array}
	\right.
	\label{eq:subs_dzia}
\end{equation}
Using~\eqref{eq:subs_dzia}, we can continue to the derivation of the \textit{distributed} iterative solution of the dual problem.

\subsection{Distributed Solution for Classification Problem}
\label{subsec:dscp}
After substituting~\eqref{eq:subs_dzia} in~\eqref{eq:alpha_eq}, we obtain the expression for $\alpha_{ik}^{\text{(prev)}}$ and $\alpha_{ik}$:
\begin{equation}
	\begin{array}{l}
		1 = y_{ik} \pmb{x}_{ik}^\intercal \left( \pmb{w} -  \alpha_{ik}^{(\text{prev})}  y_{ik}  \pmb{x}_{ik} \right) + \alpha_{ik} y_{ik}^2  \norm{\pmb{x}_{ik}}^2 \\
		+    y_{ik} {\pmb{x}_{ik}^\intercal} \left( \pmb{v}_k - \frac{1}{C_2}\alpha_{ik}^{(\text{prev})} y_{ik} {\pmb{x}_{ik}} \right) 
		+ \frac{1}{C_2} \alpha_{ik} y_{ik}^2 \norm{\pmb{x}_{ik}}^2.
	\end{array}
	\label{eq:subs_der}
\end{equation}
In~\eqref{eq:subs_der}, $\alpha_{ik}$ is an unknown variable which we express as
\begin{equation}
	\alpha_{ik} = \frac{1- y_{ik} \pmb{x}_{ik}^\intercal \pmb{w} - y_{ik} \pmb{x}_{ik}^\intercal \pmb{v}_k}{\norm{\pmb{x}_{ik}}^2 \left(1+\frac{1}{C_2}\right)} + \alpha_{ik}^{\text{(prev)}}.
	\label{eq:ait_iters}
\end{equation}
To calculate updated $\pmb{w}$ and $\pmb{v}_k$ at epoch $j$, we can apply expressions in~\eqref{eq:derres}, which yields
\begin{equation}
	\left\{
	\begin{array}{l}
		\pmb{w}_{ik}^{(j+1)}  = \pmb{w}_{ik}^{(j)} + \left(\alpha_{ik}^{(j+1)} - \alpha_{ik}^{(j)} \right) y_{ik} \pmb{x}_{ik}  ,\\
		\pmb{v}_k^{(j+1)} = \pmb{v}_k^{(j)} + \frac{1}{C_2}\left(\alpha_{ik}^{(j+1)} - \alpha_{ik}^{(j)} \right) y_{ik} \pmb{x}_{ik},
	\end{array}
	\right.
  \label{eq:cls_w_v}
\end{equation}
where $\alpha_{ik}^{(j+1)} - \alpha_{ik}^{(j)} =  \frac{1- y_{ik} \pmb{x}_{ik}^\intercal \pmb{w} - y_{ik} \pmb{x}_{ik}^\intercal \pmb{v}_k}{\norm{\pmb{x}_{ik}}^2 \left(1+\frac{1}{C_2}\right)}$ as follows from \eqref{eq:ait_iters}. As seen from the above expressions, the tasks are decoupled from each other and participants can participate in collaborative training by only sharing $\Delta \pmb w_{ik} = \pmb{w}_{ik}^{(j+1)} - \pmb{w}_{ik}^{(j)}$ at each epoch.

\begin{table}[ht]
	\centering
	\caption{Notation for Complexity Analysis}
	\begin{tabular}{lr}
		\hline
		Notation & \makecell[c]{Description} \\
		\hline
		$C_\text{add}^{d}$ & Computational cost of $d$ additions \\
		$C_\text{sub}^{d}$ & Computational cost of $d$ subtractions \\
		$C_\text{mul}^{d}$ & Computational cost of $d$ multiplications \\
		$C_\text{div}^{d}$ & Computational cost of $d$ divisions \\
		$C_{\text{sqrt}}^{d}$ & Computational cost of $d$ square roots \\
		\hline
	\end{tabular}
	\label{tab:conota}
\end{table}

\subsection{Computational Complexity of Classification}
\label{subsec:ccc}
\par Furthermore, we analyze the complexity of the straightforward implementation of our MTL-SVM classification algorithm. In this analysis, we assume that no low-level optimization steps are taken. We denote the volume of the training set as $L$, i.e., $L = n_k$ for the federated multi-task methods and $L = \sum \limits_{k=1}^{K} n_k$ in the case if all data are available at the central server. Employing the notation from~Table~\ref{tab:conota}, we arrive at the following costs $C_{\text{z}}$ of computing variable $z$:
\begin{equation}
	\left\{
		\begin{array}{l}
			C_{\pmb \alpha} = L\left(3C_\text{mul}^{d} + 2C_\text{add}^{d} + C_\text{div}^{3} + C_\text{add}^{1}\right),\\
			C_{\pmb{w}} = L\left(C_\text{mul}^{d} + C_\text{add}^{d} + C_\text{sub}^{1} + C_\text{mul}^{1}\right),\\
			C_{\pmb{v}} = L\left(C_\text{mul}^{d} + C_\text{add}^{d} + C_\text{mul}^{2} + C_\text{sub}^{1} + C_\text{div}^{1}\right),\\
			C_{\text{end}} = C_\text{sub}^{4} + C_\text{mul}^{3} + C_\text{div}^{1} + C_{\text{sqrt}}^{1}.
		\end{array}
	\right. 
\end{equation}
Here, $d$ is the dimensionality of the data. We may observe that if the computational costs of different operations are assumed to be the same, the total cost can be approximated as $C_{\text{total}}(L,d) = Ld + d + L + c$, where $c$ corresponds to the delays introduced by the code execution.

%
%

\section{Multi-task Regression}
\label{sec:mtr}
\par In this section, we formulate the dual problem for~\eqref{eq:regpr} and derive an iterative solution for individual participants in the multi-task setting.

\subsection{Dual Problem for Regression}
\label{subsec:dpr}
Using the objective function and constraints in~\eqref{eq:regpr}, we formulate the Lagrangian as
\begin{equation}
	\begin{array}{l}
		{L} (\pmb{w}, \pmb \xi^{+},  \pmb \xi^{-})
		= \frac{1}{2}\norm{\pmb{w}}^2  + \frac{C_2}{2}  \sum \limits_{i=1}^{K}\norm{\pmb{v}_k}^2
		 
		\\
		- \sum \limits_{i=1}^{K}\sum \limits^{n_k}_{i=1}\alpha_{ik}^{-} \left[ \pmb{w}^\intercal \pmb{x}_{ik} + \pmb{v}_k^\intercal \pmb{x}_{ik} -y_{ik}+\epsilon+\xi^{-}_{ik} \right]
		\\
		- \sum \limits_{i=1}^{K}\sum \limits^{n_k}_{i=1}\alpha_{ik}^{+} \left[ -\pmb{w}^\intercal \pmb{x}_{ik} - \pmb{v}_k^\intercal \pmb{x}_{ik}+y_{ik}+\epsilon+\xi^{+}_{ik} \right] 
		\\
		-\sum \limits_{i=1}^{K}\sum \limits^{n_k}_{i=1} \left(\eta^{-}_{ik}\xi^{-}_{ik} + \eta^{+}_{ik} \xi^{+}_{ik} \right) + 
		C_1\sum \limits_{i=1}^{K}  \sum \limits_{i=1}^{n_k} \left(\xi_{ik}^{+}  + \xi_{ik}^{-}  \right) 
	\end{array}
	\label{eq:reglag}
\end{equation}
and calculate its partial derivatives as follows:
\begin{equation}
	\left\{
	\begin{array}{l}
		\frac{\partial L}{\partial \pmb{w}} = \pmb{w} +\sum \limits_{k=1}^{K} \sum \limits_{i=1}^{n_k} \alpha_{ik}^{+}  \pmb{x}_{ik} -\sum \limits_{k=1}^{K} \sum \limits_{i=1}^{n_k} \alpha_{ik}^{-}  \pmb{x}_{ik}, \\
		\frac{\partial L}{\partial \pmb{v}_k} = C_2 \pmb{v}_k \! +\! \sum \limits_{i=1}^{n_k}  \alpha_{ik}^{+}  \pmb{x}_{ik} \!-\! \sum \limits_{i=1}^{n_k}  \alpha_{ik}^{-}  \pmb{x}_{ik}, \\
		\frac{\partial L}{\partial \xi_i} = C_1  -  \eta_{ik}^{+} -\alpha_{ik}^{+}.
	\end{array}
	\right.	
	\label{eq:regpd}
\end{equation}
As in the case of the classification problem, we can express $\pmb{w}$, $\pmb{v}_{k}$ and $C_2$ as
\begin{equation}
	\left\{
	\begin{array}{l}
		\pmb{w} =\sum\limits_{k=1}^{K} \sum \limits_{i=1}^{n_k} \left( \alpha_{ik}^{-} - \alpha_{ik}^{+}   \right) \pmb{x}_{ik}, \\
		\pmb{v}_k=\frac{1}{C_2}\sum \limits_{i=1}^{n_k}\left(   \alpha_{ik}^{-} -\alpha_{ik}^{+}  \right)  \pmb{x}_{ik}, \\
		C_2  =  \eta_{ik}^{+} +\alpha_{ik}^{+}, \\
		 C_1 = \eta_{ik}^{-} + \alpha_{ik}^{-}.
	\end{array}
	\right.	
	\label{eq:regexpr}
\end{equation}
We reformulate~\eqref{eq:reglag} by substituting expressions~\eqref{eq:regexpr} for $\pmb{w}$, $\pmb{v}_k$, and $C_2$. For convenience we minimize $-L(\alpha)$ given as
\begin{equation}
		\begin{array}{l}
		-L(\alpha) = \frac{1}{2} \left|\left|\sum\limits_{k=1}^{K}\sum \limits_{i=1}^{n_k}\! \left( \alpha_{ik}^{-} -  \alpha_{ik} ^{+}\right) \pmb{x}_{ik} \right|\right|^2 
		\\
		+ \frac{1}{2C_2} \sum \limits^{K}_{k=1}\left|\left|\sum\limits_{i=1}^{n_k} \left( \alpha_{ik}^{-} -\alpha_{ik}^{+} \right)  \pmb{x}_{ik} \right|\right|^2
		\\
		\!-\!  \sum \limits_{k=1}^{K} \sum \limits_{i=1}^{n_k}\left(  \alpha_{ik}^{-} \!-\!  \alpha_{ik}^{+} \right) y_{ik} 
		+ 
		\epsilon \sum \limits_{k=1}^{K} \sum \limits_{i=1}^{n_k} \left(  \alpha_{ik}^{-} + \alpha_{ik}^{+} \right).
		\end{array}
	\label{eq:lagsubs}
\end{equation}
Furthermore, we rewrite~\eqref{eq:lagsubs} as follows:
\begin{equation}
	\begin{array}{l}
		\!\!\! -L(\alpha) = \frac{1}{2} \sum \limits_{\substack{k_1=1 \\ k_2=1}}^{K} \sum \limits_{\substack{i=1 \\ j=1}}^{n_k} \left(  \alpha_{ik_1}^{-} - \alpha_{ik_1}^{+} \right) \left(  \alpha_{jk_2}^{-} - \alpha_{jk_2}^{+} \right) \pmb{x}_{ik_1}^T   \pmb{x}_{jk_2}
		\\
		+ \frac{1}{2C_2} \sum \limits^{K}_{k=1} \sum\limits_{i,j=1}^{n_k} \left( \alpha_{ik}^{-} -\alpha_{ik}^{+} \right)  \left( \alpha_{jk}^{-} -\alpha_{jk}^{+} \right)  \pmb{x}_{ik}^\intercal \pmb{x}_{jk} 
		\\
		\!-\!  \sum \limits_{k=1}^{K} \sum \limits_{i=1}^{n_k}\left(  \alpha_{ik}^{-} \!-\!  \alpha_{ik}^{+} \right) y_{ik} 
		+ 
		\epsilon \sum \limits_{k=1}^{K} \sum \limits_{i=1}^{n_k} \left(  \alpha_{ik}^{-} + \alpha_{ik}^{+} \right).
	\end{array}
	\label{eq:neglag}
\end{equation}
Therefore, we may define the dual optimization problem with respect to variables $\left(\alpha_{ik}^{+},\alpha_{ik}^{-}\right)$ as shown below:
\begin{equation}
	\begin{aligned}
		\min_{\alpha_{ik}^{+}, \alpha_{ik}^{-}} \quad & \frac{1}{2} \sum \limits_{\substack{k_1=1 \\ k_2=1}}^{K} \sum \limits_{\substack{i=1 \\ j=1}}^{n_k} \left(  \alpha_{ik_1}^{-} - \alpha_{ik_1}^{+} \right) \left(  \alpha_{jk_2}^{-} - \alpha_{jk_2}^{+} \right) \pmb{x}_{ik_1}^T   \pmb{x}_{jk_2}	\\
		\quad & + \frac{1}{2C_2} \sum \limits^{K}_{k=1} \sum\limits_{i,j=1}^{n_k} \left( \alpha_{ik}^{-} -\alpha_{ik}^{+} \right)  \left( \alpha_{jk}^{-} -\alpha_{jk}^{+} \right) \pmb{x}_{ik}^\intercal \pmb{x}_{jk} 
		\\
		\quad & \!-\!  \sum \limits_{k=1}^{K} \sum \limits_{i=1}^{n_k}\left(  \alpha_{ik}^{-} \!-\! \alpha_{ik} ^{+} \right) y_{ik} 
		+ 
		\epsilon \sum \limits_{k=1}^{K} \sum \limits_{i=1}^{n_k} \left(  \alpha_{ik}^{-} + \alpha_{ik}^{+} \right) \\
		\textrm{s.t.} \quad & 0 \leq  \alpha_{ik}^{+} \leq  C_1, k=1,...,T, i = 1,...,n_k,\\
		\quad &0 \leq  \alpha_{ik}^{-} \leq  C_1, k=1,...,T, i = 1,...,n_k.
	\end{aligned}
	\label{eq:regdp}
\end{equation}
We then employ the ADMM and minimize the objective function in~\eqref{eq:regdp} sequentially for selected $i,k$. For convenience, we express $\pmb{w}$ and $\pmb{v}_k$ via the difference $\Delta \alpha_{ik} = \alpha_{ik}^{-}-\alpha_{ik}^{+}$ as
\begin{equation}
	\left\{
	\begin{array}{l}
		\Delta  \alpha_{ik}  = \alpha_{ik}^{-}-\alpha_{ik}^{+},\\
		\pmb{w} =\sum\limits_{k=1}^{K} \sum \limits_{i=1}^{n_k} \Delta \alpha_{ik} \pmb{x}_{ik}, \\
		\pmb{v}_k =\frac{1}{C_2} \sum \limits_{i=1}^{n_k} \Delta  \alpha_{ik} \pmb{x}_{ik}.\\
	\end{array}
	\right.
	\label{eq:dalpha_reg}
\end{equation}
The impact of pair $\left(\alpha_{ik}^{+},\alpha_{ik}^{-}\right)$ can be calculated as
\begin{equation}
  \begin{array}{l}
	 \!\!\!z_i(\alpha_{ik}^{-},\alpha_{ik}^{+}) = \\ \!
	\frac{1}{2}  \Delta  \alpha_{ik}^2 ||\pmb{x}_{ik}||^2 
	\! + \! \Delta  \alpha_{ik} \pmb{x}_{ik}^\intercal \! \! \sum \limits_{\substack{k_1=1 \\
	k_1 \neq k}}^{K}\!  \left(\! \sum \limits_{\substack{j=1 \\ j \neq i}}^{n_k} \! \Delta \alpha_{jk_1} \pmb{x}_{jk_1} \!\! \right) 
	\!\!\!\\ 
	+\frac{1}{C_2}\cdot \Delta \alpha_{ik} \pmb{x}_{ik}^\intercal \left( \sum\limits_{\substack{j=1 \\ i\neq j}}^{n_k}  \Delta \alpha_{jk} \pmb{x}_{jk}  \right) + \frac{1}{2C_2} \Delta \alpha_{ik} ^2 ||\pmb{x}_{ik}||^2 
	\\
    - \Delta \alpha_{ik} y_{ik}  + \epsilon \left(\Delta  \alpha_{ik} + 2\alpha_{ik} ^{+} \right),
  \end{array}
  \label{eq:regdivco}
\end{equation}
where $ \Delta \alpha_{ik} = \alpha_{ik}^{-}-\alpha_{ik}^{+}$. 
We define the minimization problem for the quadratic function $z_i$ 
as follows:
\vspace{-0.1cm}
\begin{equation}
	  \begin{aligned}
		\min_{\Delta \alpha_{\! ik}^{ },\alpha_{\! ik}^{+}} \quad & \frac{1}{2}  \Delta  \alpha_{ik}^2 \! \norm{\pmb{x}_{ik}}^2 
		 \!+\! \Delta  \alpha_{ik}^{ } \pmb{x}_{ik}^\intercal \!\!\!\! \sum \limits_{\substack{k_1=1 \\ k_1 \neq k}}^{K} \!\!\left( 
		\! \sum \limits_{\substack{j=1 \\ j \neq i}}^{n_k} \!\! \Delta \alpha_{jk_1} \pmb{x}_{jk_1} 
		\right)	\\
		& +\frac{1}{C_2} \Delta \alpha_{ik}^{ } \pmb{x}_{ik}^\intercal \left( \sum\limits_{\substack{j=1 \\ i\neq j}}^{n_k}  \Delta \alpha_{jk} \pmb{x}_{jk}  \right) 
		\!\! + \frac{1}{2C_2} \Delta \alpha_{ik}^2 \norm{\pmb{x}_{ik}}^2 \\
        &- \Delta \alpha_{ik} y_{ik}  + \epsilon \left(\Delta  \alpha_{ik}^{ } + 2\alpha_{ik}^{+} \right) \\
        \textrm{s.t.} \quad & 0 \leq  \alpha_{ik}^{+} \leq  C_1, k=1,...,T, i = 1,...,n_k,\\
		\quad &0 \leq  \alpha_{ik}^{-} \leq  C_1, k=1,...,T, i = 1,...,n_k.
	\end{aligned}
	\label{eq:indmin}
\end{equation}
We rely on the convexity of the quadratic function $z_i(\alpha_{ik}^{-}, \alpha_{ik}^{+})$ and 
the condition $\frac{\partial z_{ik} \left( \Delta \alpha_{ik} \right)} {\partial \left( \Delta \alpha_{ik} \right)} = 0$.
\vspace{-0.3cm}
\begin{equation}
	\begin{array}{l}
	 \!\!\!	\!\!\!	\frac{\partial z_{ik}(\Delta \alpha_{ik})} {\partial \left( \Delta \alpha_{ik} \right)}\! =\! 0 \Rightarrow
		\Delta \alpha_{ik} \! \norm{\pmb{x}_{ik}}^2 
		\!+\! x_i^T \!\!\! \sum \limits_{\substack{k_1=1 \\ k_1 \neq k}}^{K} \! \left( \sum \limits_{\substack{j=1 \\ j\neq i}}^{n_k}  \! \! \! \! \Delta\alpha_{jk} \pmb{x}_{jk} \right)
		 \\
	 \!\!\!	+  \frac{1}{C_2}\cdot \pmb{x}_{ik}^\intercal \left( \sum\limits_{\substack{j=1 \\ i\neq j}}^{n_k}  \Delta \alpha_{jk} \pmb{x}_{jk}  \right) \!+ \! \frac{1}{C_2} \Delta \alpha_{ik} \! \norm{\pmb{x}_{ik}}^2
		\!- \! y_{ik} \! +\!  \epsilon\! =\! 0. \!\!\!
	\end{array}
	\label{eq:indminder}
\end{equation}
To simplify equation~\eqref{eq:indminder} with respect to $\Delta \alpha_{ik}$, we reorganize it using the following expressions:
\begin{equation}
	\left\{
	\begin{array}{l}
		\sum \limits_{\substack{j=1 \\ j\neq i}}^{n_k} \Delta  \alpha_{jk}\pmb{x}_{jk} = C_2 \pmb{v}_k -   \Delta \alpha_{ik}^{\text{(prev)}}\pmb{x}_{ik},\\
		\sum\limits_{k_1=1}^{K} \sum \limits_{j=1}^{n_k} \Delta  \alpha_{jk_1}\pmb{x}_{jk_1} = \pmb{w} - \sum \limits_{j=1}^{n_k} \Delta  \alpha_{jk}\pmb{x}_{jk}  = \pmb{w}-C_2 \pmb{v}_k.
	\end{array}
	\right.
	\label{eq:regsubs}
\end{equation}
In the next subsection, we formulate the expressions for the updated Lagrange coefficients and the hyperplane in the case of \textit{distributed} learning.

\subsection{Distributed Solution for Regression Problem}
\label{subsec:dsrp}
Similarly to the classification problem, we substitute expressions~\eqref{eq:regsubs} into~\eqref{eq:indminder}. As a result, we may derive the update for $\Delta \alpha_{ik}$ that each participant computes separately as
\begin{equation}
	\begin{array} {l}
		\Delta \alpha_{ik} = -\frac{\pmb{x}_{ik}^\intercal \left (\pmb{v}_k    - \pmb{w}\right) - y_{ik}  +  \epsilon}{ \left (\frac{1}{C_2}-1\right) \norm{\pmb{x}_{ik}}^2 } +   \Delta \alpha_{ik}^{\text{(prev)}}.\\
	\end{array}
	\label{eq:regdevup}
\end{equation}
Based on~\eqref{eq:regdevup} and~\eqref{eq:regexpr}, we define the updates for the global and local hyperplane components as
\begin{equation}
	\left\{
	\begin{array}{l}
		\Delta \pmb{w} =\sum\limits_{k=1}^{K} \sum \limits_{i=1}^{n_k}  \left(\Delta  \alpha_{ik} - \Delta \alpha_{ik}^{\text{(prev)}} \right) \pmb{x}_{ik}, \\
		\Delta \pmb{v}_k =\frac{1}{C_2} \sum \limits_{i=1}^{n_k} \left(\Delta  \alpha_{ik} - \Delta \alpha_{ik}^{\text{(prev)}} \right) \pmb{x}_{ik}.
	\end{array}
	\right.
	\label{eq:reghypup}
\end{equation}
As a result, the common component $\pmb{w}$ is computed using the information from all participants, however, without sharing the training data. The local model $\pmb{v}_k$ only requires access to the data of the corresponding participant. This supports privacy of data since instead of the complete dataset, the participants share only common components $\pmb{w}_k$, which are then combined by the coordinator into $\pmb{w}$ and propagated back to the participants.

\subsection{Computational Complexity of Regression}
\label{subsec:ccr}
\par Here, we analyze the complexity of the MTL-SVM regression using the notation from Table~\ref{tab:conota}. We arrive at the following complexity formulations:
\begin{equation}
	\left\{
	\begin{array}{l}
		C_{ \pmb\alpha^+} = C_{ \pmb\alpha^-} = L\left(2D_m + D_a + D_d + 3_a + 1_d\right),\\
		C_{\pmb{w}} = L\left(2D_m + 2D_a + 2_s\right),\\
		C_{\pmb{v}} = L\left(2D_m + 2D_a + 2_s + 2_d\right),\\
		C_{\text{end}} = 4D_s + 4D_m + 4D_a + 4_{\text{sqrt}} + 2_d + 1_m.
	\end{array}
	\right. 
	\label{eq:compan_reg}
\end{equation}
In~\eqref{eq:compan_reg}, we recompute the updated values of $\pmb{w}$ and $\pmb{v}_k$ twice due to the presence of $\alpha_{ik}^+$ and $\alpha_{ik}^-$. This effectively doubles the computational cost of the regression solution. Asymptotically, the computational cost of the regression is the same as in the case of classification.

\section{Security Analysis of Updates}
\label{sec:sau}
\subsection{Analysis of Update Structure}
\label{subsec:aus}
In the envisioned distributed system, the updates are known to the coordinator. In this subsection, we assess whether a \textit{curious server} can reconstruct the data from the observed information. According to the proposed scheme, the server has access to the public model component, $\pmb{w}$, model update of a tagged participant $k$, $\Delta\pmb{w_k}^{\left(j\right)}$ at epoch $j$, and learning parameter $C_2$. The number of points $n_k$ of the participant $k$ for the training procedure may also be known under the assumption that users share it with the coordinator. The curious server may attempt to recover the local model component $\pmb{v}_k^{(j)}$, $d \times n_k$, data matrix $\pmb{X}_k$, and vector $\pmb{y}_k$ of data labels of participant $k$. Without loss of generality, we perform further analysis for a system with a single participant $k$ and a coordinator.

Here, we use the example of the classification problem and outline the composition of updates $\Delta \pmb{w}_{k}^{\left(j+1\right)}$ and $\Delta \pmb{v}_{k}^{\left(j+1\right)}$ in~\eqref{eq:cls_w_v} for a tagged participant $k$ at epoch $j+1$ as
\begin{equation}
\left\{
 \begin{array}{l}
 \Delta \pmb{w}_{k}^{
		\left(j + 1\right)} = \sum\limits_{i=1}^{n_k}\frac{1 - y_i \left(\pmb{w}^{\left(j\right)} + \pmb{v}_{k}^{\left(j\right)}\right)^\intercal  \pmb{x}_{ik}}{\norm{\pmb{x}_{ik}}^2} \pmb{x}_{ik} y_{ik},\\
 \Delta \pmb{v}_{k}^{
		\left(j + 1\right)} = \frac{1}{C_2}\left(\sum\limits_{i=1}^{n_k}\frac{1 - y_i \left(\pmb{w}^{\left(j\right)} + \pmb{v}_{k}^{\left(j\right)}\right)^\intercal  \pmb{x}_{ik}}{\norm{\pmb{x}_{ik}}^2} \pmb{x}_{ik} y_{ik}\right).
 \end{array}
 \right.
 \label{eq:captionres}
\end{equation}
One may conclude that since $C_2$ is known to the coordinator, it can easily derive $\Delta \pmb{v}_{k}^{\left(j+1\right)}$ using $\Delta \pmb{w}_{k}^{\left(j+1\right)}$. In this case, it would be possible to reconstruct both the entire dataset $\pmb{X}_{k}$ and labels $\pmb{y}_{k}$ by solving a system of $\left(d+1\right)n_k$ non-linear equations that follow from the first expression in~\eqref{eq:captionres}. 

To restore the data precisely, the curious server should capture $\Delta\pmb{w}_k$ for each of the $(d+1)n_k$ epochs, which may significantly exceed the number of iterations. The coordinator also needs to know the number of samples used for training, $n_k$, which generally may vary among tasks. However, if both the nature of the data and at least the estimates of $n_k$ are known, the curious server can produce multiple solutions for the system of non-linear equations for the available estimates of $n_k$ and then select the most realistic answer.

\begin{figure}[ht]
	\centering
	\includegraphics[width=\linewidth]{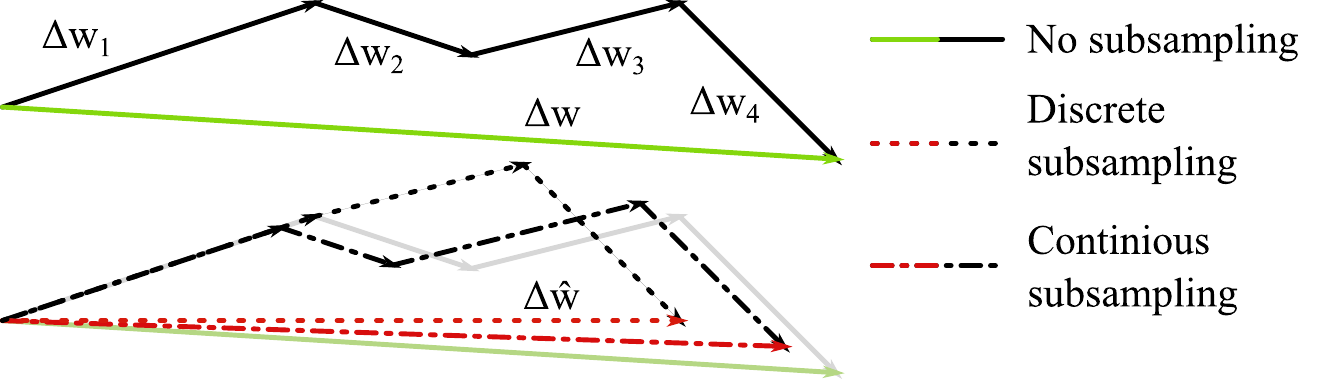}
	\caption{Example of discrete and continuous masking.}
	\label{fig:vecdist}
\end{figure}
To protect the data from reconstruction, we introduce random distortion in the transmitted updates as elaborated further in subsection~\ref{subsection:masking}. To do so, we first rewrite the expression for $\Delta \pmb{w}_{k}^{\left(j+1\right)}$ as given in~\eqref{eq:largesecurityequation} on the top of the next page.
\begin{figure*}[h!]
		\begin{equation}
		\Delta \pmb{w}_{k}^{
			\left(j + 1\right)} = \underbrace{y_1 \frac{1 - y_1 \left(\pmb{w}^{\left(j\right)} + \pmb{v}_{k}^{\left(j\right)}\right)^\intercal \begin{bmatrix}
					x_{11}\\...\\x_{1d}
			\end{bmatrix}}{\sum\limits_{i = 1}^{d}x_{1i}^2}}_{h_1} \begin{bmatrix}
			x_{11}\\...\\x_{1d}
		\end{bmatrix} + ... + \underbrace{y_{n_k} \frac{1 - y_{n_k} \left(\pmb{w}^{\left(j\right)} + \pmb{v}_{k}^{\left(j\right)}\right)^\intercal  \begin{bmatrix}
					x_{n_k 1}\\..\\x_{n_k d}
			\end{bmatrix}}{\sum\limits_{i = 1}^{d}x_{n_k i}^2}}_{h_{n_k}} \begin{bmatrix}
			x_{n_k 1}\\...\\x_{n_k d}
		\end{bmatrix}
		\label{eq:largesecurityequation}
		\end{equation}		\hrule
\end{figure*}
Let $h_{ik}^{(j)}$ denote the weight of vector $\pmb{x}_{ik}$ at epoch~$j$ for participant~$k$ in~\eqref{eq:largesecurityequation}, i.e.,
\begin{equation}
\begin{array}{c}
	 h_{ik}^{(j)} = y_{ik} \frac{1 - y_{ik} \left(\pmb{w}^{\left(j\right)} + \pmb{v}_{k}^{\left(j\right)}\right)^\intercal \pmb{x}_{ik}}{\sum\limits_{\ell = 1}^{d}x_{1\ell}^2}.
	 \end{array}
	 \label{eq:diagweight}
\end{equation}
Based on~\eqref{eq:diagweight}, the update for $\Delta \pmb{w}_{k}^{\left(j+1\right)}$ can be represented in a matrix form. If $\pmb{X}_{k}$ is a matrix of training points, and $\pmb{H}_{k}$ is a diagonal matrix of corresponding weights of the training points, then
\begin{equation}
\Delta \pmb{w}_{k}^{\left(j + 1\right)}\! =\! \pmb{X}_{k} \cdot \pmb{H}_{k}^{\left(j\right)}, \pmb{H}_{k}^{\left(j\right)}\! =\! \begin{bmatrix}
 h_{1k}^{\left(j\right)} & 0 & ... & 0 \\
 0 & h_{2k}^{\left(j\right)} & ... & 0 \\
 ... & ... & ... & ... \\
 0 & 0 & ... & h_{n_kk}^{\left(j\right)}
\end{bmatrix}\!.
\label{eq:matrixform}
\end{equation}
%


\subsection{Proposed Masking Mechanism}
\label{subsection:masking}

To prevent the curious server from recovering the data from the intercepted model updates, we apply \textit{masking} to $\Delta \hat{\pmb{w}}^{\left(j + 1\right)}$ as follows:
\begin{equation}
 \Delta \hat{\pmb{w}}_{k}^{\left(j + 1\right)} = \pmb{X}_{k} \cdot \pmb{H}_{k}^{\left(j\right)} \cdot \pmb{p}_{k}^{\left(j\right)},
\end{equation}
where $\pmb{p}_{k}^{\left(j\right)} \in \mathcal{R}^d$ is a \textit{masking vector} of random variables at iteration $j$. For \textit{discrete masking}, we utilize Bernoulli-distributed $\pmb{p}_{k}^{\left(j\right)}, p_{ik}^{\left(j\right)} \in \{0,1\}$, which, however, can be brute-forced if the number of iterations is sufficiently large. 

To make the recovery of the data even more difficult for the curious server, we propose the application of a continuous distribution for generating $\pmb{p}_{k}^{\left(j\right)}$. As an example of such distribution, we may consider the Beta distribution $\beta\left(a,b\right)$, where values are, for example, in the range $\left[0,1\right]$. 
Moreover, $\beta\left(a,b\right)$ allows the participants to adjust the shape of the distribution as desired by choosing parameters $a$ and $b$. There are two reasons behind this proposal. Firstly, if the distribution is continuous, the curious server can no longer rely on the assumption of a limited set of values and brute-force the system. Secondly, the application of a continuous distribution yields a lower performance drop compared to that of the discrete distribution since it less affects the value of the update. 

The geometrical representation of the above procedure is illustrated in~Fig.~\ref{fig:vecdist}, where the two-dimensional vector $\Delta \pmb{w}$ is represented as a sum of components $\Delta w_{i} = h_i \pmb{x}_{i}$, $i \in 1,...,n_k$. The continuous distribution of masking vector elements varies $\Delta w_{i}$, while the discrete removes selected terms from summation. 
We note that, in this case, the noise is correlated with the original data. Hence, the proposed approach differs from the one commonly used in differential privacy solutions, where the noise is not correlated with the data and is usually generated as a random variable with a Laplacian or Gaussian distribution. Our method is related to data masking that was also studied for FL~\cite{dhakal2019coded}. We present the algorithmic form of the proposed iterative classification/regression solutions with masking in Algorithm~\ref{alg:algGeneric}.

\begin{algorithm}[ht]
 \caption{MTL Classification/Regression}\label{alg:algGeneric}
 
 \begin{algorithmic}[1]
  \STATE \textbf{Input:} $\pmb{X}_{k}$, $\pmb{y}_{k}$, $C_1$, $C_2$
  \STATE \textbf{Output:} $\pmb{w}$, $\pmb{v_k}$
  \STATE Initialize 
  $\pmb{w}^{\left(0\right)}=\pmb{ 0}$
  \FOR{$k \in 1,\ldots,K$}
    \STATE $\pmb{v}_k^{\left(0\right)}=\pmb{ 0}$
  \ENDFOR
  \STATE $j=0$
  \WHILE{stopping criteria \emph{not} reached}
   \FOR{$k \in \left\{ 1,\ldots,K \right\}$}
    \STATE Initialize $\pmb{p}_{k}^{\left(j\right)}$
    \STATE Compute $\Delta \hat{\pmb{w}}_{k}^{\left(j + 1\right)} = \pmb{X}_{k} \cdot \pmb{H}_{k} \cdot \pmb{p}_{k}^{\left(j\right)}$
    \STATE Compute $\Delta \pmb{v}_{k}^{\left(j + 1\right)} = \frac{1}{C_2} \pmb{X}_{k} \cdot \pmb{H}_{k}$
    \STATE Deliver $\Delta \hat{\pmb{w}}_{k}^{\left(j + 1\right)}$ to the coordinator
   \ENDFOR
   \STATE Compute $\pmb{w}^{\left(j + 1\right)} = \sum\limits_{k=0}^{K} \Delta\hat{\pmb{w}}_{k}^{\left(j + 1\right)}$ at the coordinator
   \STATE Disseminate $\pmb{w}^{\left(j + 1\right)}$ among the participants
  \ENDWHILE
 \end{algorithmic}
\end{algorithm}

For participant $k, k \in \left\{ 1,\ldots,K \right\}$, the algorithm above takes data $\pmb{X}_k$, labels $\pmb{y}_k$, learning parameters $\left( C_1, C_2 \right)$. The outputs of the algorithm are participant-specific $\pmb{v}_k$ and common $\pmb{w}$.

\section{Numerical Results}
\label{sec:numres}
In this section, we provide selected numerical results and qualitative conclusions on the behavior of the proposed algorithm for solving classification/regression problems in the presence of stragglers. We also study the impact of different masking mechanisms on the resulting performance. Finally, we briefly explore the security of the system in the presence of a curious server.

We measure the performance of the following three methods utilized in both regression and classification problems:
\begin{itemize}
	\item \textit{Global SVM} -- The data of all participants are collected by the coordinator, which trains one global model $\pmb{w}$ and shares it. Importantly, in this case, all of the data are disclosed to the coordinator, which also significantly increases the load on the network and violates privacy requirements.
	\item \textit{Local SVM} -- Each participant trains a local model, $\pmb{w}_k$, without using any external communications. This method provides perfect privacy as no data leaves the participant. However, it results in lower accuracy of the trained model since the volume of training data available locally is smaller than in the case of the global method.
	\item \textit{MTL-SVM} -- The participant trains a model consisting of two components. Global model $\pmb{w}_k$ of the participant $k$ is computed locally, and then the update $\Delta \pmb{w}_k$ is shared with the coordinator. Local model $\pmb{v}_k$ is computed and kept locally. The coordinator processes global model updates received from the participants and sends back the new global model. The training is performed iteratively and requires communication between the coordinator and the participants. In this method, the participants do not explicitly share their data with the coordinator, yet a global model is trained. This allows the coordinator to disseminate the global model to newly joined participants, instead of recomputing it from scratch. The load on the network depends only on the number of participating nodes and the dimensionality of the model.
\end{itemize}
To summarize, the global SVM method trains a single model by aggregating and processing all training data from the participants. Local SVM does not disclose any of the data, and the node is limited to its own training set. 
In MTL-SVM, participants send global model updates, which are aggregated at the coordinator to produce the common global model, which is 
personalized by the local model. 
\vspace*{-0.5cm}

\subsection{Simulation Parameters and Datasets}
\label{subsec:simparms}
In this subsection, we describe our scenario, outline the structure of the experiments, and define the employed system parameters. Numerical results are obtained and processed with a custom simulation tool in MATLAB. We implement the algorithm in our simulator for both classification and regression problems. 


Our setup comprises one coordinator and $T=20$ participants, each with its own task. We study the performance of the proposed classification algorithm utilizing the commonly used UCIHAR dataset for human activity recognition~\cite{10.1007/978-3-642-35395-6_30,anguita2013public}. The dataset contains data collected from multiple sensors during different activities such as sitting, walking, etc. The ``sitting'' class is separated from all other classes. Each of the data points contains $d = 561$ frequency and time domain features extracted from raw sensor measurements. The set contains data for multiple individuals, which are considered to be different participants in our experiment. For regression, we construct synthetic datasets, where we control the mean and variance of $\pmb x_{ik}$ for each participant separately, which allows us to introduce tasks with dissimilar data. In this case, learning a singular global model might be infeasible due to statistical differences between the feature characteristics of different participants. 
The main system parameters are summarized in Table~\ref{tab:ml_parm}. 

We structure our numerical experiments as follows. We begin with a scenario where all participants have equal computing capabilities. 
Computing delays for a single participant are generated as random variables drawn from a Gaussian distribution since concurrent processes might affect the overall time of a single epoch. The mean and variance of the distribution are proportional to the complexity and, therefore, depend on $d$ and $n_k$. The parameters of the dependency of the mean and standard deviation on $d$ and $n_k$ are fitted based on the delay measurements performed on the computer. For the global method, we assume that the coordinator has the same computing capacity as all of the participants combined. 

\begin{table}[ht]
	\centering
	\caption{Key System Parameters}
	\label{tab:ml_parm}
	\begin{tabular}{lr}
		\hline
		Parameter & \makecell[c]{Value} \\
		\hline
		Datasets & \makecell[r]{UCIHAR,\\ Synthetic}\\
		Number of participants & 20 \\
		Tasks per participant & 1 \\
		Computing delay model & Gaussian \\
		Cross validations & 5 \\
		Training sub-set size & 70\% of all samples \\
		Participant hardware & \makecell[r]{Heterogeneous,\\ Homogeneous} \\
		\hline
	\end{tabular}
\end{table}

Furthermore, we model the \textit{diverse} computing capabilities of the participants. 
In this case, delays of a selected participant are obtained by dividing the corresponding random variables drawn from a Gaussian distribution by a constant factor that corresponds to its hardware capabilities. 
The employed modeling approach reflects \textit{hardware heterogeneity} in real systems, where different equipment (e.g., smartphones, smart glasses, notebooks, etc.) is utilized, which may lead to participants being unable to deliver their updates. 
In our particular example, we monotonically increase the hardware capabilities factor value from $1.0$ to $10.0$. 
In addition to hardware heterogeneity, we control the \textit{data heterogeneity} of the synthetic dataset by varying the statistical properties of the training data. 
%
For clarity, the core structure of our numerical results is provided in Table~\ref{tab:my_label}.
\begin{table}[h!]
    \centering
    \caption{Scenario Characteristics in Numerical Results}
    \label{tab:my_label}
    \begin{tabular}{lcccccc}
        \hline
        \multirow{3}{*}{} & \multicolumn{3}{c}{Classification} & \multicolumn{3}{c}{Regression} \\
        \hline
         & 7.2.1 & 7.2.2 & 7.2.3 & 7.3.1 & 7.3.2 & 7.3.3 \\
        \hline
        Hom. data & \checkmark & \checkmark & \checkmark & \checkmark & & \\
        Het. data & & & & & \checkmark & \checkmark \\
        Hom. HW & \checkmark & \checkmark & & \checkmark & \checkmark & \\
        Het. HW & & & \checkmark & & & \checkmark \\
        \hline
    \end{tabular}
\end{table}

To assess the performance of our regression algorithm, we employ the R-squared coefficient, which is commonly used to evaluate the goodness of regression fit and is defined for samples from task $k$ as
%
	$R^2 \!=\! 1 -\! \frac{\sum_{i=1}^{n_k} \left(y_{ik} - a(\pmb{x}_{ik})\right)^2}{\sum_{i=1}^{n_k}\left(y_{ik} \!- \frac{1}{n}\!\sum_{j=1}^{n_k}y_{jk}\right)^{\!2}} .$
In the case of classification, we measure true positive (TP) and true negative (TN) values and compare the results in terms of balanced accuracy $\frac{\text{TP} + \text{TN}}{2}$. 
%
%
\subsection{Evaluation of Classification Algorithm}
\subsubsection{Equal Computing Capacities, No Stragglers}
\label{subsec:cls_eco}
We consider a scenario where all nodes are responsive and have equal computational capacities. We measure the time required to complete a single epoch and compare it with that of the state-of-the-art MOCHA algorithm~\cite{smith2017federated}, which is built on a similar principle and aims at tackling the problem of stragglers. 

\begin{figure}[h!]
	\centering
	\includegraphics[width=\linewidth]{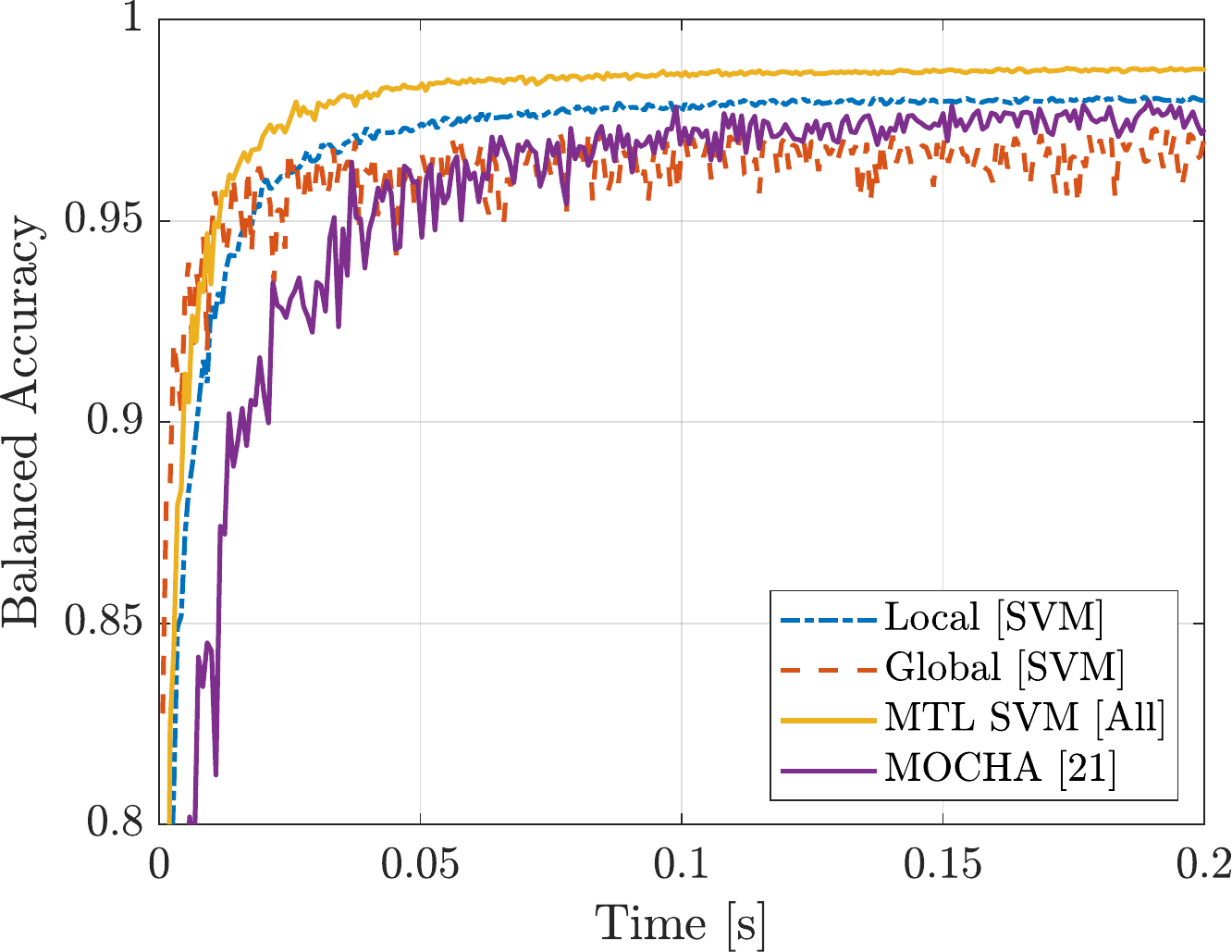}
	\caption{Classification, UCIHAR: no stragglers.}
	\label{fig:evsvssvm_st}
\end{figure}

In Fig.~\ref{fig:evsvssvm_st}, we provide the comparison in terms of balanced accuracy. The coordinator, in this particular case, waits for all of the participants to deliver an update. These settings are denoted by the label \textit{[All]} in the legend here and in subsequent figures. We observe that MTL-SVM outperforms other considered methods. In particular, the global method results in the lowest performance, while the local method demonstrates comparable values. We note that the local method can be outperformed by multi-task solutions if the dataset volume is limited. In this case, the multi-task algorithms allow us to leverage the information on the data of other participants. 
In addition, MTL-SVM rapidly converges due to the optimization of one Lagrange multiplier at each iteration in the solution of the dual problem. This could prove beneficial for devices with limited computing capacities and energy constraints as they can perform the calculations and reenter the power-saving mode.
\par
To analyze the computing delays, we select the optimal regularization constants for both methods and conduct the training procedure 64 times using the Windows release of MATLAB R2021a on a stock Lenovo P330 workstation with an i7-8700 CPU. We observe that MTL-SVM trains the models for 20 tasks in $4.47$\,s, with a standard deviation of $0.031$\,s, while the state-of-the-art multi-task algorithm requires $5.28$\,s with a standard deviation of $0.068$\,s, which represents a $20\%$ improvement in speed.
\subsubsection{Equal Computing Capacities, Stragglers}
Further, we study the impact of the presence of stragglers on the performance of our classifier. We assume equal computing capabilities for all participants, which means the delays are drawn from the same distribution. In contrast to the previous experiment, 
the coordinator has a waiting period $T_{wait}$, during which the participants are expected to complete calculating their updates. If during $T_{wait}$, some of the participants fail to deliver $\Delta \pmb{w}_i^{\left(j\right)}$, they are considered to be stragglers at epoch $j$. The aggregated global update $\Delta \pmb{w}^{\left(j\right)}$ is then calculated based on the individual updates that are received successfully. The global update, $\Delta \pmb{w}^{\left(j\right)}$, is sent to all participants, including stragglers. 

The coordinator can adjust the parameter $T_{wait}$. For example, participants may evaluate their performance with a common benchmark and store maximum epoch completion time. The coordinator acquires these data when participants join and select, e.g., the maximum from the list of received values. 
We explore four cases with different waiting periods $T_{wait}$. 
The corresponding percentages of successfully responded participants are given in the legend in the brackets as, e.g., \textit{[45\%]}.

\begin{figure}[h!]
	\centering
	\includegraphics[width=\linewidth]{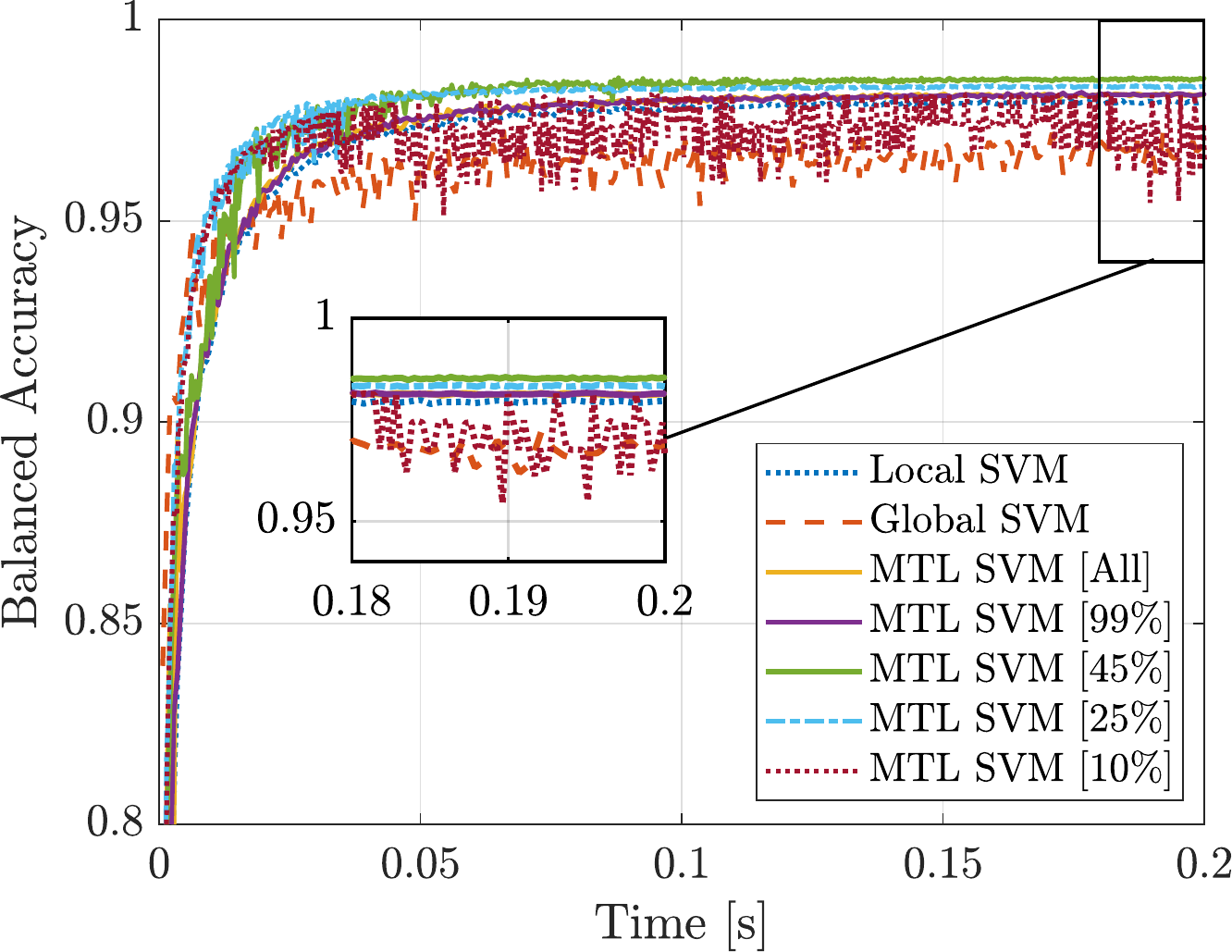}
	\caption{Classification, UCIHAR: homogeneous hardware.}
	\label{fig:cstrset}
\end{figure}

In Fig.~\ref{fig:cstrset}, we explore the impact of different percentages of responded participants on the achieved performance in terms of convergence and the final balanced accuracy. 
The presence of stragglers accelerates the convergence at the beginning due to the fact that the waiting period, $T_{wait}$, is lower than the epoch duration of the \textit{[All]} configuration. This enables a faster exchange of updates, thus accelerating the learning process. However, this effect is limited by the amount of information utilized by the non-straggling participants. For example, as illustrated in~\ref{fig:cstrset}, the $10\%$ response ratio yields poor stability due to an insufficient amount of information delivered for the construction of the model. In this particular case, the balanced accuracy degrades to the values, comparable to the ones achieved in the case of learning a single global model.
\hspace{-0.1cm}
\subsubsection{Diverse Computing Capacities, Stragglers}

\begin{figure}[h!]
	\centering
	\includegraphics[width=\linewidth]{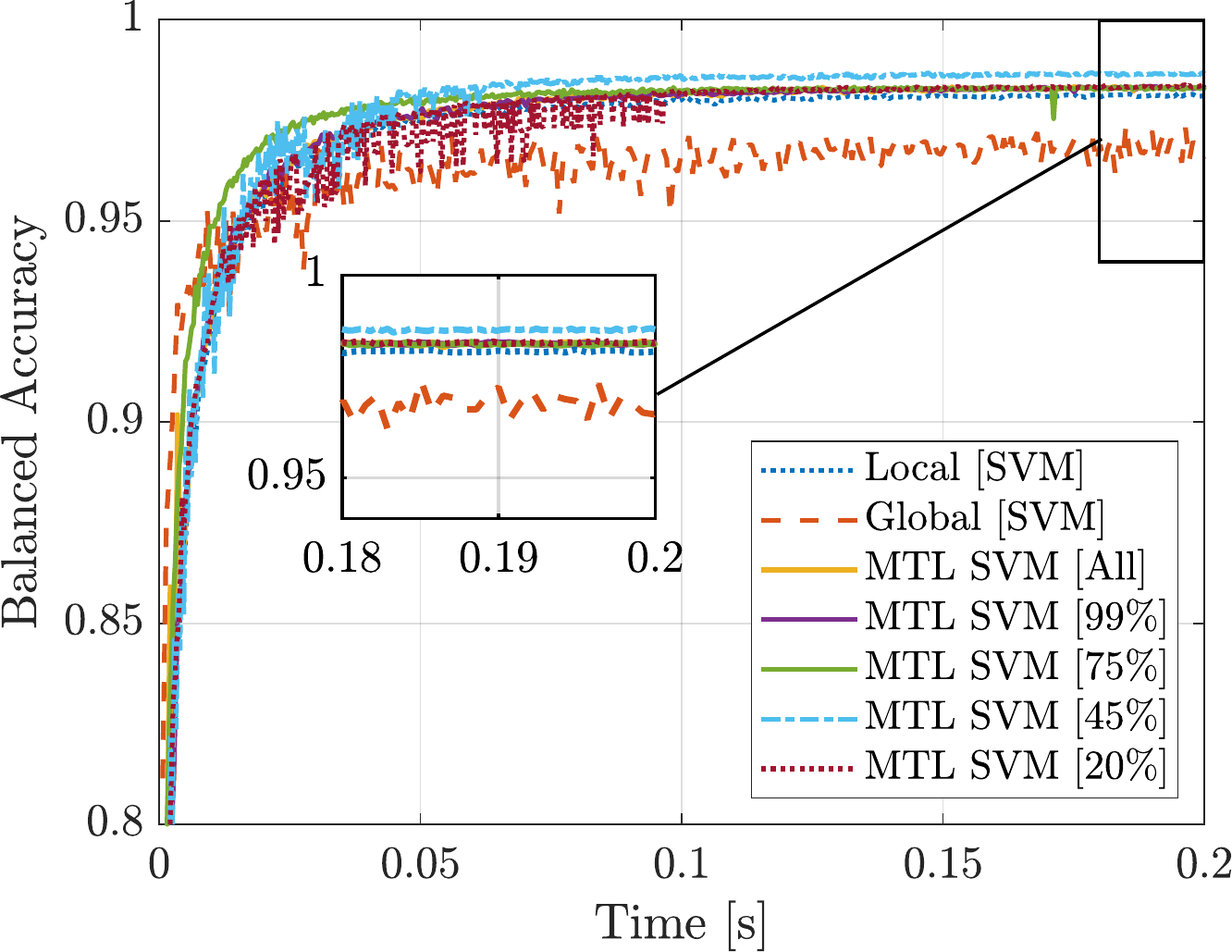}
	\caption{Classification, UCIHAR: heterogeneous hardware.}
	\label{fig:cstrset_het}
\end{figure}

In addition, we explore the impact of introducing heterogeneous hardware, where delays are modeled using the hardware capabilities factor. 
In Fig.~\ref{fig:cstrset_het}, we may observe that the final balanced accuracy is slightly higher than that in the case of homogeneous stragglers. The results also reveal that if the waiting period leads to $20\%$ of the participants responding on time, convergence toward the solution is accelerated, albeit at the cost of reduced numerical stability, compared to other cases. The degradation in numerical stability can be observed in the form of sporadic drops in balanced accuracy. 
Although some of the participants 
fail to submit the common component $\Delta\pmb{w}_k^{\left(j\right)}$ to the coordinator on time at epoch $j$, they still receive the global common component, $\Delta\pmb{w}^{\left(j\right)}$. This component, however, does not include the information from these straggling participants, thus making the model less suitable for them if the data are heterogeneous. For higher percentages of responsive participants, the balanced accuracy increases, as well as the numerical stability of the method improves. The final value of the balanced accuracy reaches similar values in all cases.
\subsection{Evaluation of Regression Algorithm}
\subsubsection{Equal Computing Capacities, Homogeneous Data}
\label{sss:reg_homdhomh}
\begin{figure}[h!]
	\centering
	\includegraphics[width=\linewidth]{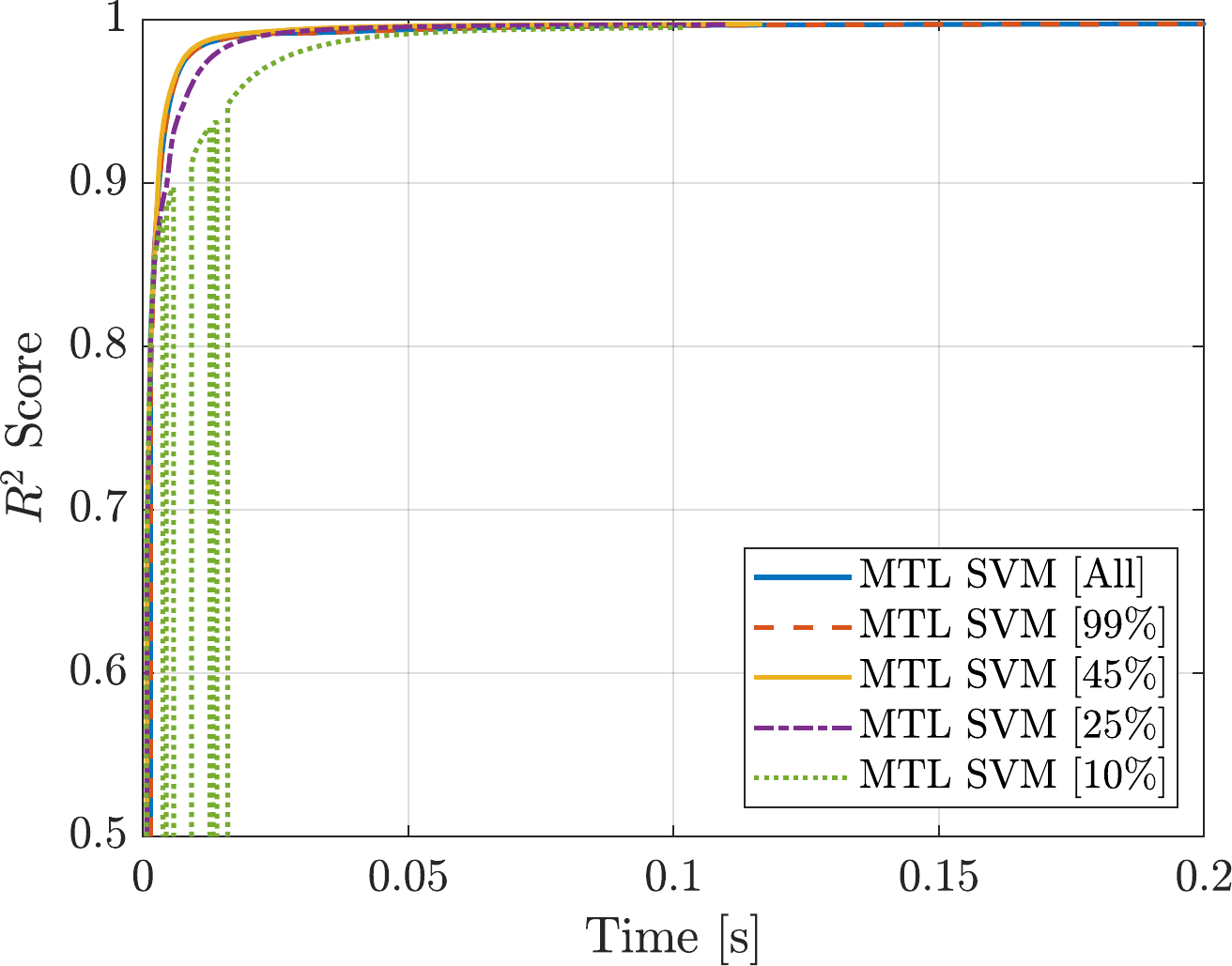}
	\caption{Regression, synthetic: homogeneous node hardware, identical data distribution.}
	\label{fig:rstrset}
\end{figure}

We begin our assessment of the regression algorithm with the case of equal computing capabilities and identical distribution of data across participants. In Fig.~\ref{fig:rstrset}, we evaluate the performance in terms of $R^2$
using a synthetic dataset with $T = 20$ tasks, each with $n_k = 200$ samples and $d = 200$ features. The noise added to the synthetic data represents measurement noise and is adjusted to maintain the SNR of $20$\,dB. We may notice that the MTL-SVM method with waiting period $T_{wait}$ corresponding to the $45\%$ of participants responding results in slightly faster convergence in absolute time due to shorter epochs. As we decrease $T_{wait}$ further and the coordinator receives fewer responses, the convergence rate degrades due to insufficient information from the participants. In summary, a reduction of $T_{wait}$ may yield marginal improvements in the case of a fully homogeneous scenario.
If we explore the results for smaller values of $T_{wait}$, we observe that for $25\%$ of participants responding, the convergence rate decreases. The case where $T_{wait}$ is reduced further and only $10\%$ of participants deliver updates on time, results in numerical instability. However, in all of the cases, the method converges to a stable solution with a high $R^2$ level.
\subsubsection{Equal Computing Capacities, Diverse Data}
\label{sss:homhhetd}
While our previous results in~\ref{sss:reg_homdhomh} assume that the data of the participants are homogeneous, in reality, this may not hold, and each participant requires model personalization. We, therefore, also investigate the behavior of the proposed methods in a scenario of heterogeneous data on nodes with homogeneous hardware.
\begin{figure}[h!]
	\centering
	\includegraphics[width=\linewidth]{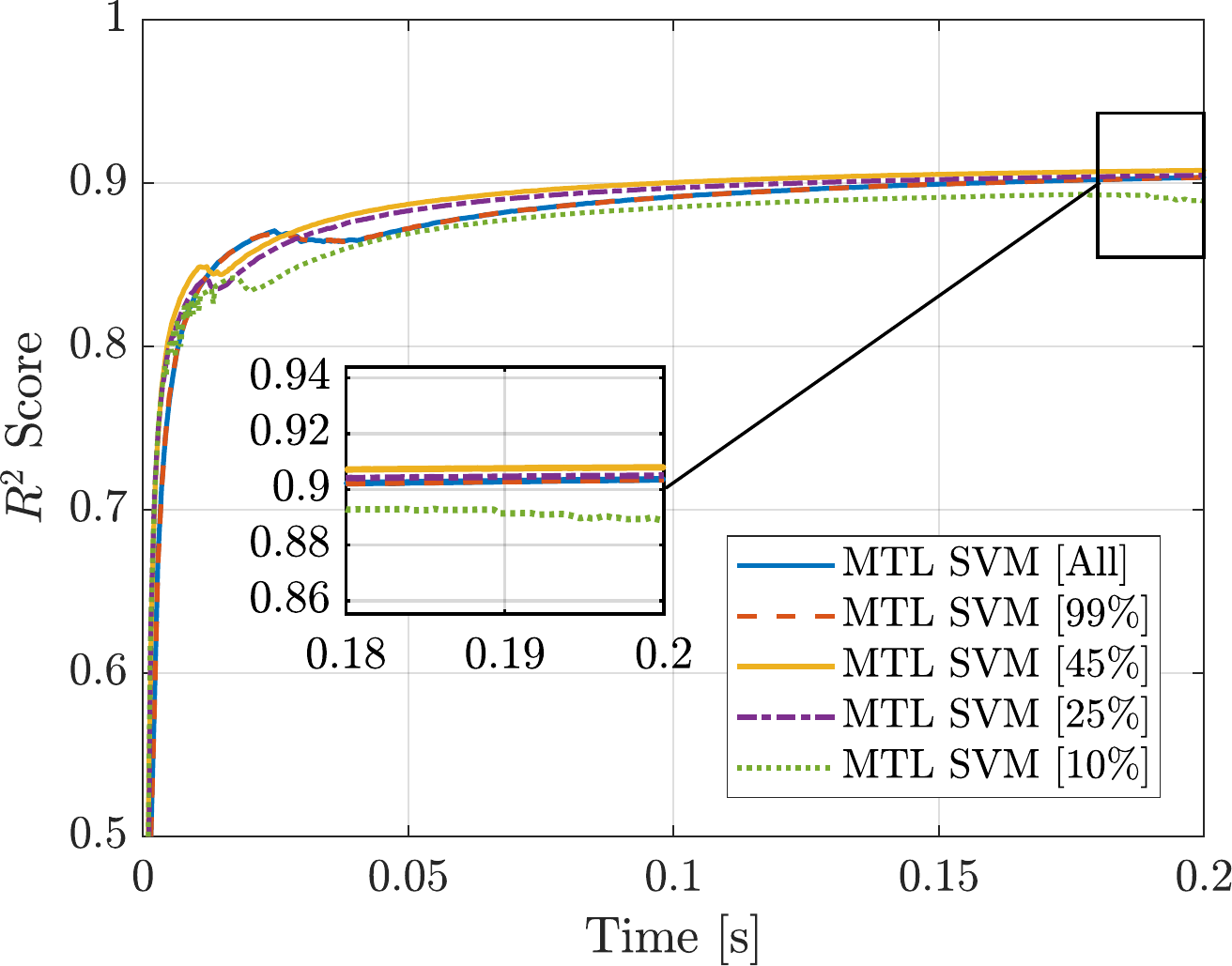}
	\caption{Regression, synthetic: heterogeneous data and homogeneous node hardware.}
	\label{fig:hetreg}
\end{figure}
In Fig.~\ref{fig:hetreg}, we observe several effects caused by the heterogeneity of the data across participants. Compared to Fig.~\ref{fig:rstrset}, the convergence rate increases noticeably for the cases where $T_{wait}$ is configured to allow $45\%$ and $25\%$ of participants to respond on time. Despite a drop in the $R^2$ metric, for both $45\%$ and $25\%$ response percentages, the method still converges at a higher rate than in the case when the coordinator waits for all of the participants. For the $10\%$ case, the convergence degrades compared to other cases. Moreover, we observe that at the later epochs, the $R^2$ metric declines slowly. This effect is caused by the insufficient amount of information delivered with $\Delta \pmb{w}$. In this particular scenario, we observe the system balancing between having enough data delivered by the participants and reducing the waiting period to shorten the epoch time. The overall performance is, however, degraded due to the heterogeneity of the data. 
\subsubsection{Diverse Computing Capacities, Diverse Data}
\begin{figure}[h!]
	\centering
	\includegraphics[width=\linewidth]{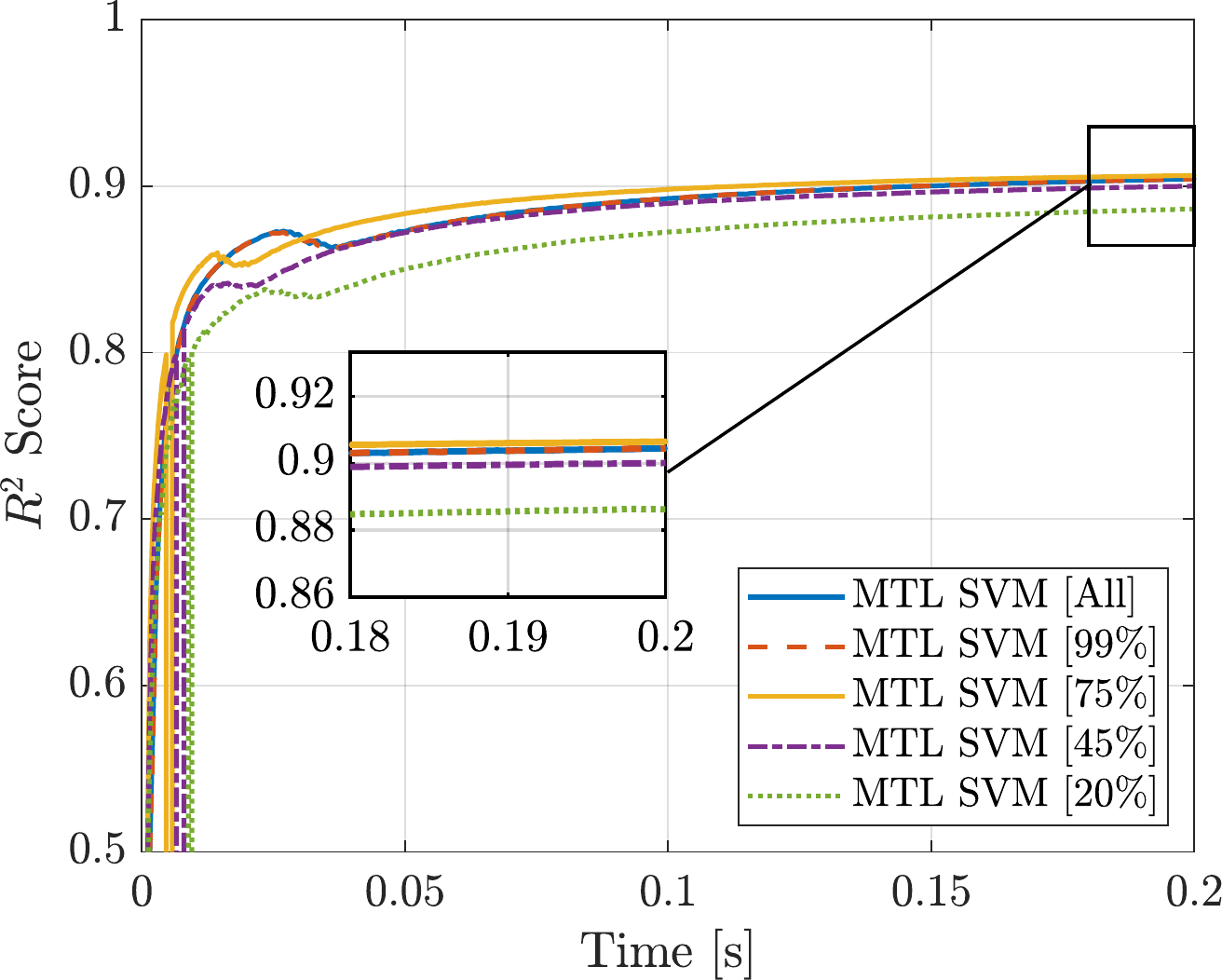}
	\caption{Regression, synthetic: heterogeneous node hardware and data.}
	\label{fig:rstrset100}
\end{figure}

In Fig.~\ref{fig:rstrset100}, we present the results where the data and computing power of the participating nodes are heterogeneous. 
We may notice that setting $T_{wait}$ for allowing $45\%$ percent of participants to respond is no longer feasible and provides no benefit in terms of learning speed. If $75\%$ of participants respond on time, it still provides a visible improvement in terms of convergence rate, while $20\%$ percent of participants responding on time yield noticeable performance degradation. In this scenario, we observe effects from both the reduction of $T_{wait}$ and increased heterogeneity of the stragglers. This results in less information being delivered from the participants and, thus, a sharper decrease in the quality of the model for lower percentages of data delivered on time.
The performance of the method with both the hardware and data being heterogeneous does not change significantly compared to the scenario in~\ref{sss:homhhetd}.
\subsection{Security Aspects: Impact of Masking}
We evaluate the impact of masking by considering two types of distributions of elements of the masking vector, $p$, namely, Bernoulli and $\beta\left(a, b\right)$ distributions. We also employ truncated UCIHAR with $n_k=100$ samples per task, which is marked as ``light UCIHAR''. With this dataset, it is possible to illustrate the benefit of the proposed MT-SVM algorithm for data-deficient scenarios. For the Bernoulli distribution, the parameter is adjusted in the range of $\left\{1, 0.75, 0.5, 0.25\right\}$, for $\beta\left(a,b\right)$ distribution, we set $a=2, b=0.5$ (the mean of the $\pmb{p}$ then equals $\frac{a}{a+b} = 0.8$) to avoid complete drop-out of the data points. Since $\beta\left(a,b\right)$ distribution can affect all data points, we control the ratio of the points where we do not apply masking within the range $\left\{1, 0.75, 0.5, 0.25\right\}$.
\begin{figure}[h!]
 \centering
 \begin{subfigure}{0.48\linewidth}
  \includegraphics[width=\linewidth]{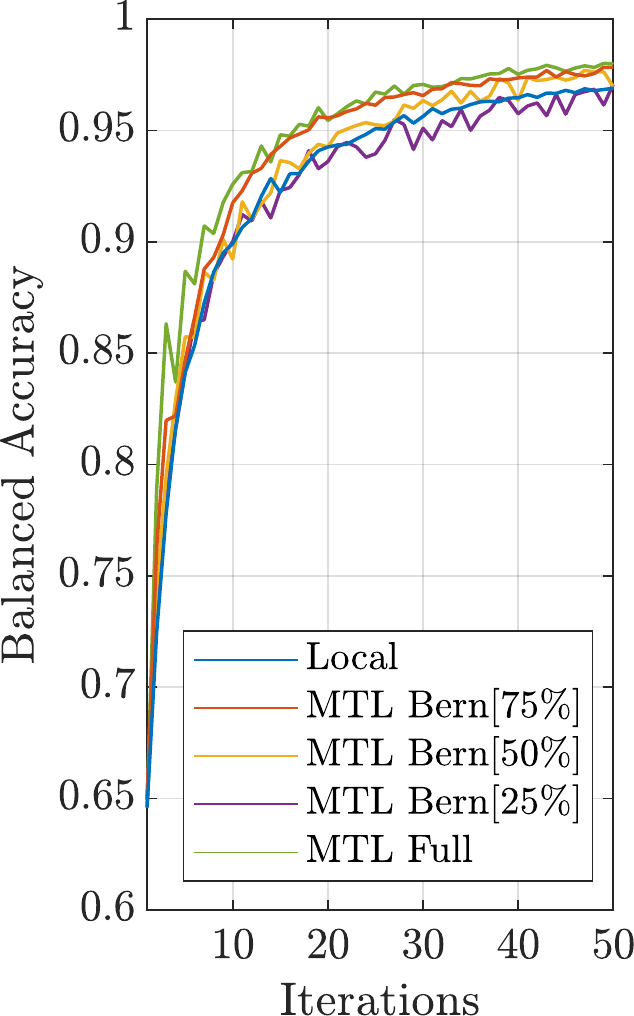}
  \subcaption{Original UCIHAR dataset.}
 \end{subfigure}
 \begin{subfigure}{0.48\linewidth}
  \includegraphics[width=\linewidth]{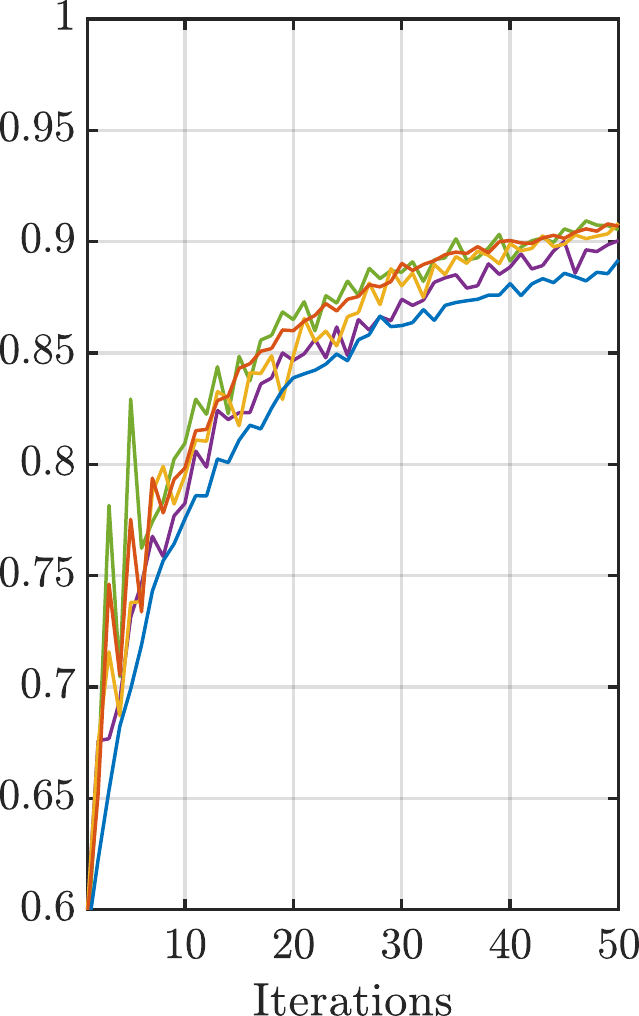}
  \subcaption{Light UCIHAR dataset.}
 \end{subfigure}
 \caption{Classification, UCIHAR: effects of Bernoulli masking. The percentage in the brackets corresponds to the fraction of training points that are affected by the masking procedure.}
 \label{fig:berncode}
\end{figure}
For the Bernoulli masking, in Fig.~\ref{fig:berncode}, we may observe a significant performance degradation for low values of distribution parameters, i.e., for $E[p] = 0.25$, since most of the points are excluded from the training procedure. For parameter values of $\left\{1, 0.75, 0.5\right\}$, the degradation is not as severe, albeit the difference with MTL-SVM without masking is noticeable. For a truncated dataset with $n_k = 100$, the performance impact is more evident. This dataset also highlights one of the strong sides of the MTL, when participants hold a limited volume of data. Even for the highest exclusion ratio of $0.25$, the resulting balanced accuracy is greater than that in the case of the local mode of operation.
\begin{figure}[h!]
 \centering
 \begin{subfigure}{0.48\linewidth}
  \includegraphics[width=\linewidth]{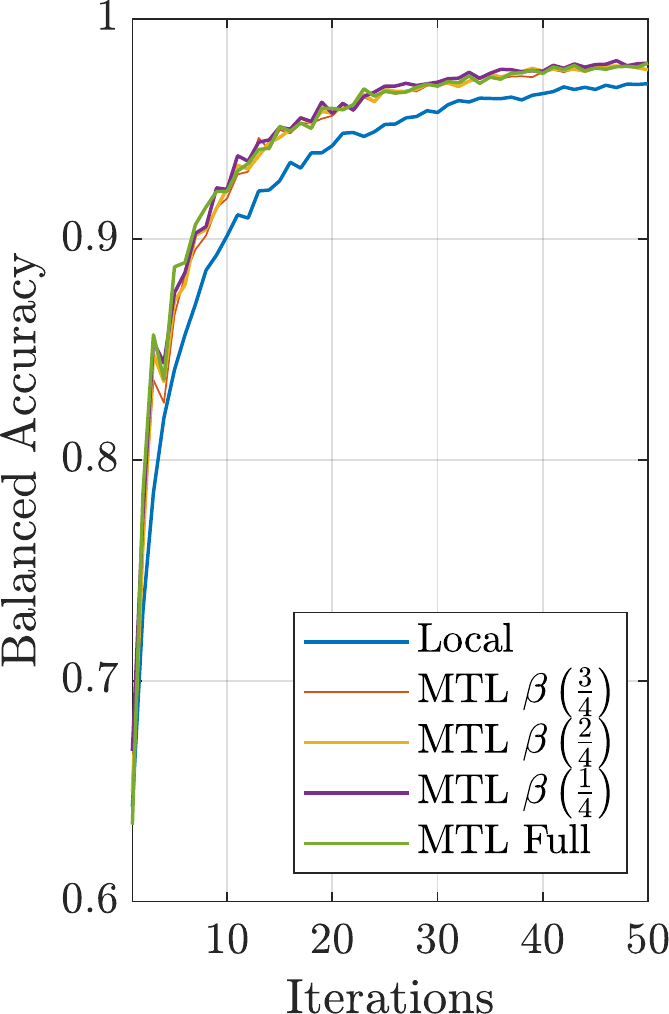}
  \subcaption{Original UCIHAR dataset.}
 \end{subfigure}
 \begin{subfigure}{0.48\linewidth}
  \includegraphics[width=\linewidth]{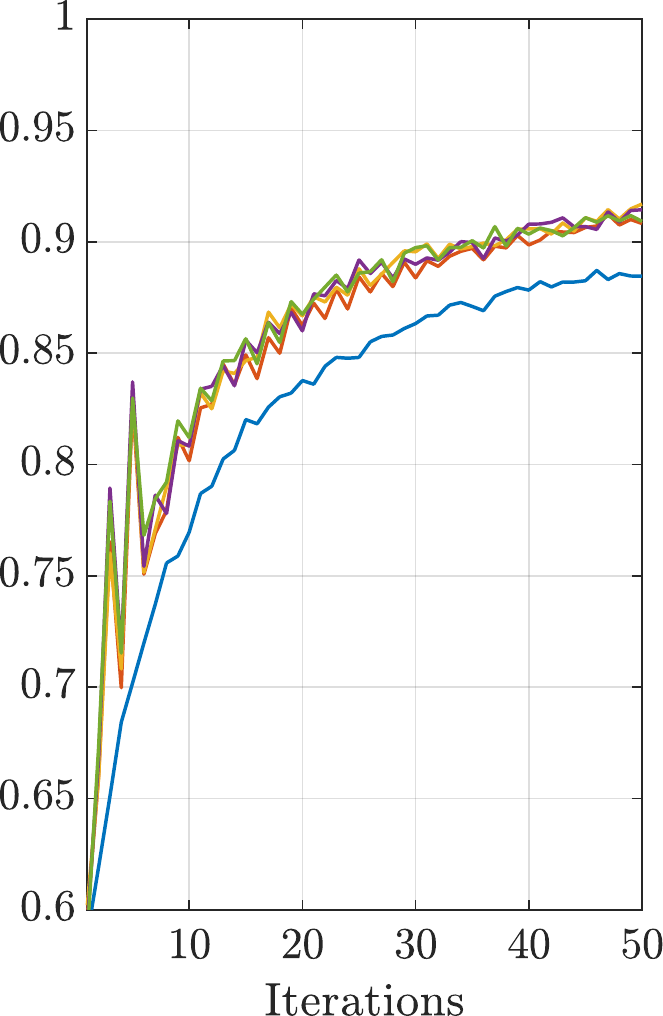}
  \subcaption{Light UCIHAR dataset.}
 \end{subfigure}
 \caption{Classification, UCIHAR: effects of $\beta\left(2,0.5\right)$ masking on balanced accuracy for original and truncated versions of the dataset. In $\beta\left(r\right)$, $r$ corresponds to the ratio of the points not affected by the masking.}
\end{figure}
The application of the $\beta\left(a,b\right)$ distribution yields a more stable performance, compared to the Bernoulli distribution. The difference between applying these two distributions is such that the Bernoulli distribution only excludes random data points from participation in a single epoch. Contrary to this, the $\beta\left(a,b\right)$ distribution decreases the contribution of every data point by a random value. Thus, the effect of randomization based on the $\beta\left(a,b\right)$ distribution is less noticeable than that of the case of the Bernoulli distribution, which results in more stable performance. We may observe that with the $\beta\left(a,b\right)$ distribution, the overall performance is nearly unaffected by the introduced masking. In the case of the light UCIHAR dataset, we can observe that the MTL method consistently outperforms the local method.

\section{Conclusions}
\label{sec:concl}
In this work, we study a distributed learning system deployed over a network of heterogeneous participants and provide an algorithmic solution for the federated multi-task classification and regression problems. The proposed algorithm allows personalizing the learning model for each participant without sharing the training data and improves the performance, compared to that of the locally trained models provide. The method is especially beneficial in the case of the low volumes of data available to individual participants.
Based on the results of the numerical performance evaluation, we come to the following conclusions regarding the designed algorithm.
\par
The proposed iterative distributed solution provides performance comparable to or superior to learning on locally available data. In the case of learning a single global model on heterogeneous data aggregated at the coordinator, the advantage of our distributed iterative method is even more pronounced. 
In all the considered cases, the proposed MTL-SVM method outperforms the global method by a noticeable margin. The performance of our distributed method also exceeds the performance of local learning methods. Such an advantage is especially noticeable in scenarios where participants hold an insufficient volume of data. 
When compared with other similar methods, our method requires 20\% less time for one epoch.
This is especially important for participants with limited computing power or energy source. 
\par The method is robust to the presence of stragglers, and its performance may also be boosted by adjusting the waiting period $T_{wait}$ to a smaller value. This could decrease the epoch duration at the cost of missing updates from slower participants.
In the presence of stragglers, our proposed method demonstrates stable convergence and performance even in those cases, where only a quarter of the participants are responding on time. This holds for both homogeneous and diverse computing capacities. 
\par
Furthermore, in the case of the low training data volume, participants can reach a higher accuracy than in the case of learning only local data. This may aid participants that are limited in terms of available data either due to its scarcity or low volume of available storage. This is also beneficial for participants with poor hardware since they may not be able to store all of the data at the same time.
Finally, the proposed masking mechanisms allow participants to prevent data disclosure in case of message interception or a curious server threat model. The theoretical analysis of privacy guarantees is one of the future directions we intend to explore.

\bibliographystyle{IEEEtran}
\bibliography{refs_mtl}
\end{document}